\documentclass[lettersize,journal]{IEEEtran}
\usepackage{amsmath,amsfonts}
\usepackage{algorithmic}
\usepackage{algorithm}
\usepackage{array}
\usepackage[caption=false,font=normalsize,labelfont=sf,textfont=sf]{subfig}
\usepackage{textcomp}
\usepackage{stfloats}
\usepackage{url}
\usepackage{verbatim}
\usepackage{graphicx}
\usepackage{cite}
\usepackage{booktabs}
\usepackage{wrapfig}
\usepackage[table,dvipsnames]{xcolor}
\usepackage{multicol}
\usepackage{tikz}
\usepackage{hyperref}
\usepackage[normalem]{ulem}
\usepackage{soul}
\usepackage{multirow}
\usepackage{amsmath,bm}
\usepackage{pifont}

\usepackage{orcidlink}
  
\begin{document}

\title{Fed-HeLLo: Efficient Federated Foundation Model Fine-Tuning with Heterogeneous LoRA Allocation}


\author{Zikai Zhang\orcidlink{0009-0006-7437-6424},
Ping Liu\orcidlink{0000-0002-3170-3783},
Jiahao Xu\orcidlink{0000-0002-4888-252X},
Rui Hu*\orcidlink{0000-0003-3317-1765}
\thanks{Zikai Zhang, Ping Liu, Jiahao Xu, Rui Hu are with the Department of Computer Science and Engineering, University of Nevada, Reno, Reno, NV, 89557, USA (e-mail: zikaiz@unr.edu; pingl@unr.edu; jiahaox@unr.edu; ruihu@unr.edu).}
\thanks{*Rui Hu is the Corresponding Author: ruihu@unr.edu}
}

\markboth{Journal of \LaTeX\ Class Files,~Vol.~14, No.~8, June~2025}%
{Shell \MakeLowercase{\textit{et al.}}: A Sample Article Using IEEEtran.cls for IEEE Journals}


\maketitle

\begin{abstract}
Federated Learning (FL) has recently been utilized to collaboratively fine-tune foundation models (FMs) across multiple clients. Notably, federated low-rank adaptation (LoRA)-based fine-tuning methods have recently gained attention, which allows clients to fine-tune FMs with a small portion of trainable parameters locally. However, most existing methods do not account for the heterogeneous resources of clients or lack an effective local training strategy to maximize global fine-tuning performance under limited resources. In this work, we propose {Fed-HeLLo}, a novel federated LoRA-based fine-tuning framework that enables clients to collaboratively fine-tune an FM with different local trainable LoRA layers. To ensure its effectiveness, we develop several heterogeneous LoRA allocation (HLA) strategies that adaptively allocate local trainable LoRA layers based on clients' resource capabilities and the layer importance. 
Specifically, based on the dynamic layer importance, we design a Fisher Information Matrix score-based HLA (FIM-HLA) that leverages dynamic gradient norm information. To better stabilize the training process, we consider the intrinsic importance of LoRA layers and design a Geometrically-Defined HLA (GD-HLA) strategy. It shapes the collective distribution of trainable LoRA layers into specific geometric patterns, such as \textit{Triangle}, \textit{Inverted Triangle}, \textit{Bottleneck}, and \textit{Uniform}. Moreover, we extend GD-HLA into a randomized version, named Randomized Geometrically-Defined HLA (RGD-HLA), for enhanced model accuracy with randomness. By co-designing the proposed HLA strategies, we incorporate both the dynamic and intrinsic layer importance into the design of our HLA strategy.
To thoroughly evaluate our approach, we simulate various complex federated LoRA-based fine-tuning settings using five datasets and three levels of data distributions ranging from IID to extreme Non-IID. The experimental results demonstrate the effectiveness and efficiency of Fed-HeLLo with the proposed HLA strategies. The code is available at \url{https://github.com/TNI-playground/Fed_HeLLo}.
\end{abstract}


\section{Introduction}
\IEEEPARstart{F}{oundation} models (FMs)~\cite{dosovitskiy2020image, devlin2019bert, lyu2023attention, study2022, zhou2023comprehensive}, characterized by their extensive parameter counts ranging into millions or billions, serve as robust initial weights for a variety of downstream tasks~\cite{wu2023visual,wang2018glue} via fine-tuning. However, employing FMs presents substantial challenges, especially the high computational costs of fine-tuning the model.

To mitigate the high computational requirement of fine-tuning FMs, researchers have developed a variety of parameter-efficient fine-tuning (PEFT) methods. These methods either selectively update a subset of model parameters or incorporate a small set of trainable layers, such as in prompt tuning~\cite{lester2021power}, low-rank adaptation (LoRA)~\cite{hu2021lora}, and BitFit~\cite{zaken2022bitfit}.
%

Recent works~\cite{zhao2023fedprompt,guo2023promptfl,zhang2023towards,babakniya2023slora,sun2024improving} have started to explore the combination of FL and various PEFT methods due to their lightweight training structures. 
{Federated Learning (FL)~\cite{konecny2016federated,zawad2025fedcust} is a distributed machine learning framework that enables multiple clients to collaboratively update a global model.}
%
%
%
In the FL process, each client trains a local model on its private dataset and sends model updates to a central server for aggregation. Incorporating PEFT techniques in this process improves parameter efficiency, reducing both computational and communication overhead.
In the literature,
FedPrompt~\cite{zhao2023fedprompt} and PromptFL~\cite{guo2023promptfl} combine FL with prompt tuning, updating only the soft prompts to improve parameter and communication efficiency. Due to the advantages of LoRA in handling complex downstream tasks with mid-size dataset~\cite{zhang2023fedpetuning}, particularly those involving significant domain shifts, several studies have explored federated LoRA-based fine-tuning methods~\cite{zhang2023towards,babakniya2023slora}. 
{For instance, FedIT~\cite{zhang2023towards} combines FL with LoRA for federated instruction-tuning. SLoRA~\cite{babakniya2023slora} introduces a novel initialization method for LoRA layers to improve the convergence speed of LoRA in FL.}

\begin{figure}[t]
\centering
\includegraphics[width=0.45\textwidth]{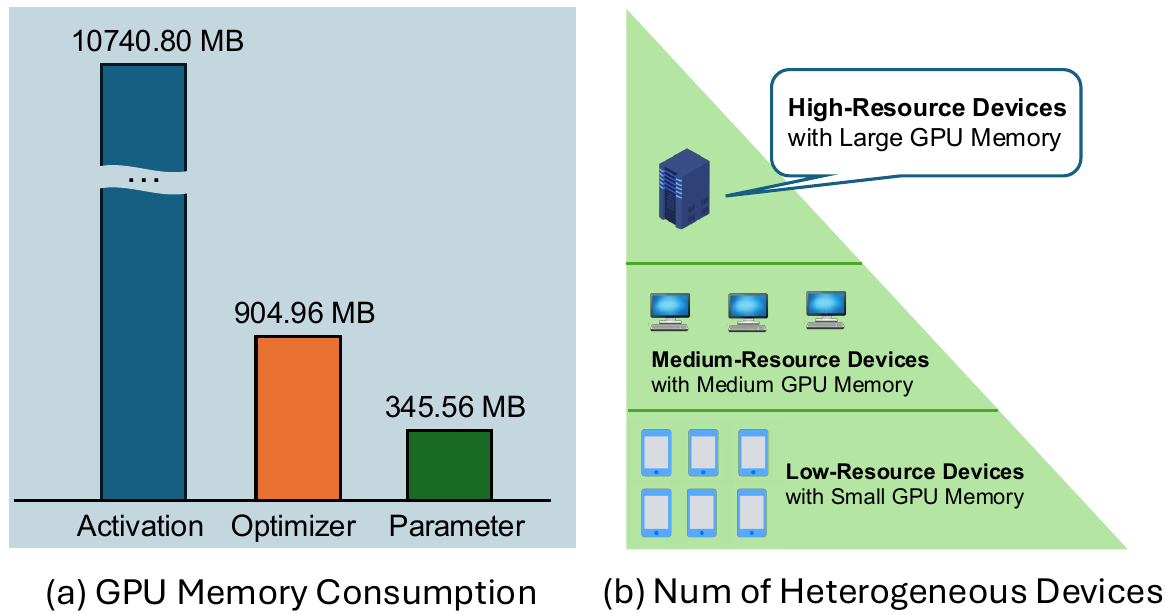}
\caption{(a) GPU memory usage for the ViT-base model, which includes a 12-layer encoder with LoRA applied to each layer. Batch size is 128.
(b) The distribution of heterogeneous devices in a practical scenario.
}
\label{fig:motivation}
\end{figure}

Despite advancements in federated LoRA-based fine-tuning, the persistent challenges are the memory bottlenecks experienced by clients with limited resources.
While LoRA is designed to be parameter-efficient, it still requires substantial memory usage
during training, similar to full-model fine-tuning~\cite{li2023fedtp}. 
As shown in Figure~\ref{fig:motivation} (a), the memory usage for training a ViT-base~\cite{krizhevsky2009learning} model
including the memory for model parameters, optimizer states, and activations. The model parameters include the weights and biases of the frozen pre-trained model as well as trainable parameters.
The optimizer states store gradients and necessary momentum estimates for trainable parameters.
Activations are intermediate results in forward process that are stored for computing gradients in backpropagation. In this LoRA-based fine-tuning case, model parameters take only 3\% of the memory, while optimizer states and activations together take 97\% of the memory.

%
Furthermore, as shown in Figure~\ref{fig:motivation} (b), clients in real-world FL have heterogeneous memory capabilities, and the majority of them have low-resource devices with limited memory capabilities.
This presents substantial challenges to the deployment of federated LoRA-based fine-tuning, as clients with limited memory resources are not capable of completing the local training task effectively and efficiently. 
%
%
%
%
In this case, FL usually assigns local training tasks only to high-capability clients (known as exclusive learning~\cite{kim2022depthfl}), which results in a significant loss of data from low-capability clients. Otherwise, FL conducts local training based on the resources of the least powerful client (known as straggler learning~\cite{mitliagkas2016asynchrony}). This is sub-optimal because it wastes the resources of high-capability clients that could perform more computation. 

To address the memory constraint issue in heterogeneous resource settings,
existing works~\cite{su2023fedra,cho2023heterogeneous,wang2024flora,bai2024flexlora} employ training strategies where resource-limited clients use LoRA with smaller ranks~\cite{cho2023heterogeneous,wang2024flora,bai2024flexlora} or update
only partial
LoRA layers~\cite{su2023fedra}. 
Specifically, methods that use smaller LoRA ranks primarily reduce the memory cost for optimizer states, thus the reduction is not significant.
Other approaches that train only partial LoRA layers can lower both activation and optimizer memory costs. However, they allocate trainable LoRA layers randomly without considering layer importance.
%
%
%

In this work, we strategically allocate partial LoRA layers to reduce memory requirements for local training in resource-constrained clients.
Specifically, we propose a novel \textbf{Fed}erated LoRA-based fine-tuning framework with \textbf{He}terogeneous \textbf{L}oRA al\textbf{Lo}cation, called \textbf{Fed-HeLLo}. 
This framework allows clients with different local trainable LoRA layers to collaboratively fine-tune an FM. 
However, it remains challenging to ensure that their local model updates can effectively contribute to obtain a well-trained global model. 
To address this, we design heterogeneous LoRA allocation (HLA) strategies that consider the varying importance of different LoRA layers.
Our contributions are summarized as follows: 
\begin{itemize}
\item {To address the memory constraint in heterogeneous clients, we propose a novel federated LoRA-based fine-tuning framework called Fed-HeLLo.} 
This method enables clients to collaboratively fine-tune an FM by updating different local LoRA layers in each training round. 
To the best of our knowledge, this is the first work to design a federated LoRA-based fine-tuning framework with heterogeneous LoRA allocation. 



\item We design a Fisher Information Matrix score-based HLA (FIM-HLA) that leverages dynamic gradient norm information to allocate trainable LoRA layers on the server.

\item To better stabilize the training process, we design a Geometrically-Defined HLA (GD-HLA) strategy by considering the intrinsic importance of different layers, such as the relationships among shallow, middle, and deep feature extraction. This strategy allocates trainable LoRA layers to different clients so that the collective distribution of LoRA layers forms a specific geometrical pattern, such as \textit{Triangle} ($\rhd$), \textit{Inverted Triangle} ($\lhd$), \textit{Bottleneck} ($\bowtie$) and \textit{Uniform} ($\approx$). We also extend GD-HLA to a randomized version, Randomized Geometrically-Defined HLA (RGD-HLA) for enhanced robustness. By co-designing these HLA strategies, we incorporate both the dynamic and intrinsic layer importance into our framework. 


\item To thoroughly evaluate our approach, we simulate various federated LoRA-based fine-tuning settings using diverse datasets and varying data distributions for both visual and language tasks. The experimental results demonstrate the effectiveness and efficiency of our framework. 
\end{itemize}

\section{Related Works}\label{sec:relatedwork}

\textbf{Parameter-efficient fine-tuning.} 
Resource-constrained devices cannot train an FM from scratch or fully fine-tune an FM due to the numerous trainable parameters.
Consequently, researchers have turned to {PEFT} techniques~\cite{ding2022delta, lester2021power, zaken2022bitfit, hu2021lora}, which involve updating only a small subset of the model's parameters. It can be mainly categorized into two types: additive and selective PEFT methods. Additive PEFT methods introduce new trainable layers to the model while keeping the pre-trained model weights frozen, enabling efficient task-specific fine-tuning. For example, prompt tuning~\cite{lester2021power} adds task-specific soft prompts, which are then fine-tuned to adapt the model to new tasks. 
LoRA~\cite{hu2021lora} introduces trainable low-rank matrices that are added to the existing weights of the pre-trained model for fine-tuning. 
Selective PEFT methods update only a subset of the pre-trained model's parameters. For instance, BitFit~\cite{zaken2022bitfit} updates only the bias parameters, while LN-Norm~\cite{basu2024strong} updates the layer normalization parameters in ViT models. Further advancements have been made with hybrid approaches that combine multiple PEFT techniques to enhance fine-tuning efficiency~\cite{he2021towards, zhou2023autopeft}. Among these PEFT methods, LoRA has demonstrated more stable and effective performance for complex downstream tasks with mid-size datasets~\cite{ding2022delta, zhang2023fedpetuning}. Without loss of generality, we focus on LoRA in this work.

{\textbf{Federated LoRA-based fine-tuning.} 
Given the resource challenges posed by full-model fine-tuning~\cite{li2023fedtp} in FL, federated LoRA-based fine-tuning has gained increasing attention~\cite{liu2023differentially,zhang2023towards,babakniya2023slora,wu2024fedfmsl,wu2024fedbiot,sun2024improving,gao2025flowertune}. These approaches improve the parameter efficiency and reduce computational and communication costs~\cite{yu2023federated, zhuang2023foundation}.
%
%
For example, FedIT~\cite{zhang2023towards} proposes a vanilla federated fine-tuning method with LoRA and establishes benchmarks for PEFT in FL.
SLoRA~\cite{babakniya2023slora} designs a two-stage fine-tuning method that combines full model and LoRA-based fine-tuning to establish a better initialization for LoRA weights and improve the global model accuracy. 
FedFMSL~\cite{wu2024fedfmsl} proposes a two-stage LoRA fine-tuning framework. In the first stage, all LoRA layers are trainable to quickly adapt to the current task. In the second stage, the model is initialized with the weights obtained from the first stage, and a Sparsely Activated LoRA (SAL) algorithm is introduced to progressively activate LoRA layers for saving computational costs.
FedBiOT~\cite{wu2024fedbiot} proposes a federated LoRA-based fine-tuning algorithm that uses compressed pre-trained weights (named as emulator) on each client. By using an emulator that simulates the behavior of the original FMs on client, the computational cost for local training can be reduced.
DP-LoRA~\cite{liu2023differentially} is a vanilla approach that incorporates a Gaussian-based differential privacy (DP) mechanism into federated LoRA-based fine-tuning. By adding noise to model updates, it protects clients' privacy during the FL process.
FFA-LoRA~\cite{sun2024improving} introduces a LoRA layer initialization strategy to improve model accuracy and communication efficiency. By default, the LoRA weight matrix $A$ (input projection) is initialized with small random values, and matrix $B$ (output projection) is initialized as zero. They prove that fine-tuning only matrix $B$ is sufficient for efficient model fine-tuning.
However, these approaches often fail to account for the heterogeneity in clients' resources.}

{\textbf{FL with heterogeneous clients.}
To address the heterogeneous resource capabilities of clients in FL, existing methods~\cite{diao2020heterofl,horvath2021fjord,kim2022depthfl,liu2022no} can be broadly categorized into two types: width-wise scaling and depth-wise scaling. Width-wise scaling approaches~\cite{diao2020heterofl, horvath2021fjord} prune model channels to create a smaller and more streamlined local model for clients with limited resources to update. Depth-wise scaling approaches~\cite{kim2022depthfl, liu2022no} assign only the first few layers of the model for resource-limited clients to update, thus reducing resource requirements for clients. However, these approaches focus on full-model training and cannot be directly applied to LoRA-based fine-tuning.
There are several existing works~\cite{cho2023heterogeneous,wang2024flora,bai2024flexlora,su2023fedra,zhang2024fed} in federated LoRA-based fine-tuning consider clients with heterogeneous resources. 
HETLoRA~\cite{cho2023heterogeneous} employs a heterogeneous LoRA rank strategy, allowing clients to train different numbers of LoRA ranks based on their memory capacity, with a hierarchical aggregation rule where shared ranks are averaged across multiple clients, while higher ranks are aggregated among fewer clients.
FLoRA~\cite{wang2024flora} is designed to mitigate aggregation noise during federated LoRA-based fine-tuning. It introduces a stacking strategy for aggregation, where LoRA layer updates from clients are stacked along the rank dimension respectively, and then used to reconstruct the full weight matrices. For each next round, the LoRA layers are initialized on each clients and only the full weight matrices are updated.
FlexLoRA~\cite{bai2024flexlora} introduces a novel aggregation strategy that first aggregates LoRA weight matrices $A$ and $B$ separately. Then those aggregated LoRA weights are used to reconstruct the full weight matrices for each LoRA layer. Finally, they 
apply singular value decomposition (SVD) to redistribute the LoRA weight matrices among clients with varying ranks.
However, reducing LoRA ranks in these rank-based methods can only reduce the memory cost of optimizer states, making them less effective in memory-constrained environments.
FedRA~\cite{su2023fedra} introduces a layer-wise LoRA allocation strategy that randomly selects partial LoRA layers in each round to address resource constraints among clients. However, this method overlooks the varying importance of different layers in the global model’s accuracy.}

{In Table~\ref{tab:survey}, we provide an overall comparison of our approach with the state-of-the-art (SOTA) methods discussed above. By evaluating four key aspects: whether to use PEFT techniques, whether to consider resource and model heterogeneity, and whether incorporating a dynamic training strategy based on training-time model performance. Overall, our method demonstrates clear advantages over existing works.}

\begin{table}[htbp]
  \centering
  \caption{{Compared with SOTA Works.}}
  \scalebox{0.80}{
    \begin{tabular}{ccccccccc}
    \toprule[2pt]
    \textbf{{Method}}   & \textbf{{PEFT}} & \textbf{{Resource Hete.}} & \textbf{{Model Hete.}} & \textbf{{Dynamic Training}}  \\
    \midrule
    {FedTP~\cite{li2023fedtp}}                         & {\ding{55}}        &{\ding{55}}         &{\ding{55}}         &{\ding{51}}                  \\
    {FedIT~\cite{zhang2023towards}}               & {\ding{51}}        & {\ding{55}}        & {\ding{55}}         & {\ding{55}}                   \\
    {SLoRA~\cite{babakniya2023slora}}                      & {\ding{51}}        & {\ding{55}}        & {\ding{55}}        &  {\ding{55}}                \\
    {FedFMSL~\cite{wu2024fedfmsl}}                      &{\ding{51}}         & {\ding{55}}        & {\ding{55}}        &  {\ding{51}}                 \\
    {FedBiOT~\cite{wu2024fedbiot}}                     &{\ding{51}}         & {\ding{55}}        &{\ding{55}}         & {\ding{51}}                 \\
    {DP-LoRA~\cite{liu2023differentially}}                      & {\ding{51}}        & {\ding{55}}        & {\ding{55}}        & {\ding{55}}                \\
    {FFA-LoRA~\cite{sun2024improving}}                      & {\ding{51}}        & {\ding{55}}        & {\ding{55}}        & {\ding{55}}                  \\
    {FLoRA~\cite{wang2024flora}}                          & {\ding{51}}        & {\ding{51}}        & {\ding{51}}        & {\ding{55}}               \\
    {FlexLoRA~\cite{bai2024flexlora}}                         & {\ding{51}}        & {\ding{51}}        & {\ding{51}}        & {\ding{55}}               \\
    {HETLoRA~\cite{cho2023heterogeneous}}                   & {\ding{51}}        & {\ding{51}}        & {\ding{51}}        &  {\ding{55}}              \\
    {FedRA~\cite{su2023fedra}}                          &{\ding{51}}         & {\ding{51}}        & {\ding{51}}        & {\ding{55}}                 \\
    \midrule
    \textbf{{Ours}}                   & {\ding{51}} & {\ding{51}} & {\ding{51}} & {\ding{51}}  \\
    \bottomrule[2pt]
    \end{tabular}%
    }
  \label{tab:survey}%
  \arrayrulecolor{black}
\end{table}%

\section{Our Proposed Approach: Fed-HeLLo}
In this section, we present Fed-HeLLo, a novel framework of federated LoRA-based fine-tuning. 
Fed-HeLLo is designed for training heterogeneous LoRA layers across different clients. 
To improve the global model's performance in Fed-HeLLo, we design the HLA strategy that adaptively allocates trainable LoRA layers according to both the dynamic and intrinsic importance. 

\subsection{FedHeLLo: Federated LoRA-based Fine-Tuning with Heterogeneous LoRA Allocation}\label{sec:problem-formulation} 
The idea behind LoRA is to constrain the update of a pre-trained weight matrix $\boldsymbol{W}^0 \in \mathbb{R}^{d \times k}$ by a low-rank decomposition $\boldsymbol{W}^0 + \Delta \boldsymbol{W} = \boldsymbol{W}^0 + \boldsymbol{B} \boldsymbol{A}$, where $\boldsymbol{B} \in \mathbb{R}^{d \times r}$, $\boldsymbol{A} \in \mathbb{R}^{r \times k}$, and the rank $r \ll \min(d, k)$. During training, $\boldsymbol{W}^0$ remains frozen, and only $\boldsymbol{A}$ and $\boldsymbol{B}$ are updated. We consider a pre-trained model $\boldsymbol{\Phi}^0 = \{\boldsymbol{W}_{(1)}^0, \cdots, \boldsymbol{W}_{(l)}^0, \boldsymbol{W}_{rest}^0\}$ with $l$ transformer layers, where each transformer layer is $\boldsymbol{W}_{(j)}^0$ with $j \in [1, l]$, and $\boldsymbol{W}_{rest}^0$ represents the remaining layers. LoRA is typically applied to specific modules within transformer layers, such as the attention weights~\cite{hu2021lora}. 
{We denote the LoRA layer added to the $j$-th transformer layer as 
$\boldsymbol{\theta}_{(j)} = \{\boldsymbol{A}_{(j)}, \boldsymbol{B}_{(j)}\}$
and any additional trainable parameters in the embedding or classifier layers as $\Delta\boldsymbol{W}_{rest}$.} 
Here, we define the set of trainable parameters as { \(\boldsymbol{\mathcal{\theta}}:= \{\boldsymbol{\mathcal{\theta}}_{(1)}, \cdots, \boldsymbol{A}_{(l)}, \boldsymbol{B}_{(l)}\}\), simplifying the representation by omitting \(\Delta\boldsymbol{W}_{rest}\).}

We consider a federated LoRA-based fine-tuning system where $n$ clients collaboratively fine-tune a global model $\boldsymbol{\Phi}$, consisting of frozen weights $\boldsymbol{\Phi}^0$ and trainable parameters $\boldsymbol{\mathcal{\theta}}$, under the coordination of a central server. Each client $i\in [n]$ holds a local dataset $D_i\sim\mathcal{P}_i$, where $\mathcal{P}_i$ represents the data distribution specific to client $i$.
The problem of federated LoRA-based fine-tuning can be formulated as follows:
\begin{equation}\label{eq:ffm_obj}
\min_{\boldsymbol{\mathcal{\theta}}} \mathcal{L}(\boldsymbol{\mathcal{\theta}}) := \frac{1}{n}\sum_{i=1}^{n} \mathcal{L}_{i}(\boldsymbol{\mathcal{\theta}}; \boldsymbol{\Phi}^0),
\end{equation}
where $\mathcal{L}_{i}(\boldsymbol{\mathcal{\theta}}; \boldsymbol{\Phi}^0) := \mathbb{E}_{x \in  D_i}[\ell(\boldsymbol{\mathcal{\theta}};\boldsymbol{\Phi}^0,x)]$ denotes the local loss of client $i$, and $\ell(\boldsymbol{\mathcal{\theta}};\boldsymbol{\Phi}^0,x)$ is the loss function of the model $\boldsymbol{\Phi}=\{\boldsymbol{\mathcal{\theta}}, \boldsymbol{\Phi}^0 \}$ for a data point $x$ sampled from ${D}_i$. 
%
To solve this problem, one can follow the classic FL algorithm FedAvg~\cite{mcmahan2017communication}. Specifically, before training, the server broadcasts the initial global model $\boldsymbol{\Phi}_g^0=\{\boldsymbol{\mathcal{\theta}}_g^0,\boldsymbol{\Phi}^0\}$ to all clients, where $\boldsymbol{\mathcal{\theta}}_g^0$ represents the initialized global trainable parameters. Clients then save this model as their initial local models $\boldsymbol{\Phi}_i^0, i\in[n]$. In each training round $t\in[T]$, the server sends its current global trainable parameters $\boldsymbol{\mathcal{\theta}}_g^t$ to a subset of $s$ clients, denoted by $\mathcal{S}^t$. Each selected client initializes its local trainable parameters $\boldsymbol{\mathcal{\theta}}_i^{t}$ with $\boldsymbol{\mathcal{\theta}}_g^t$ and obtains the local model $\boldsymbol{\Phi}_i^{t}= \{\boldsymbol{\mathcal{\theta}}_i^{t},\boldsymbol{\Phi}^0\}$. Based on $\boldsymbol{\Phi}_i^{t}$, the client updates $\boldsymbol{\mathcal{\theta}}_i^{t}$ individually using $D_i$ to obtain $\boldsymbol{\mathcal{\theta}}_i^{t+1}$ and sends its local update $ \boldsymbol{\delta}_i^{t+1} := \boldsymbol{\mathcal{\theta}}_i^{t+1} - \boldsymbol{\mathcal{\theta}}_i^t$ to the server. The global trainable parameters are then updated as follows:
$
\boldsymbol{\mathcal{\theta}}_g^{t+1} := \boldsymbol{\mathcal{\theta}}_g^t + ({1}/{|\mathcal{S}^t|})\sum_{i\in \mathcal{S}^t}\boldsymbol{\delta}_i^{t+1},
$
resulting in the global model $\boldsymbol{\Phi}_g^{t+1}=\{\boldsymbol{\mathcal{\theta}}_g^{t+1},\boldsymbol{\Phi}^0\}$ for the next round of training.

\begin{figure}[t]
\centering
\includegraphics[width=0.50\textwidth]{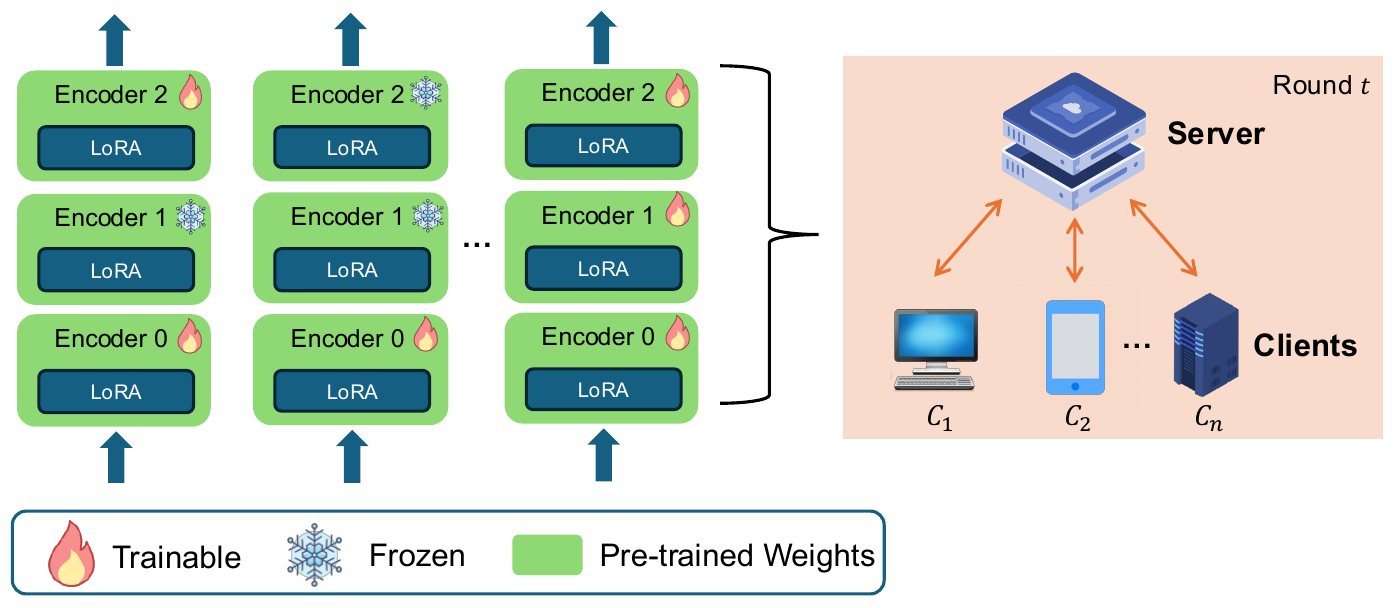}
\caption{Overview of the Fed-HeLLo framework.
}
\label{fig:framework}
\end{figure}
In the naive FedAvg process described above, clients should update all the trainable parameters in $\boldsymbol{\mathcal{\theta}}_g$ locally, which includes $l$ LoRA layers. This requires substantial on-device GPU memory for optimizer states and intermediate activations (as shown in Figure~\ref{fig:motivation}). For the optimizer state, it increases linearly with the size of the trainable parameters. When applying LoRA to each transformer layer, all the activation of these layers need to be stored for backpropagation, {akin} to full-model fine-tuning. 
This high memory requirement makes the local training task infeasible for clients with limited resources~\cite{zhang2023memoryadaptive}. 

Inspired by the recent model freezing methods proposed in \cite{su2023fedra,ardakani2024slimfit,pan2024lisa}, freezing a model layer can eliminate the storage of its dynamic activations, significantly reducing memory requirements for local training in federated fine-tuning. 
Moreover, training partial layers reduces the storage of optimizer states.
With model freezing, resource-constrained clients can train only partial LoRA layers based on their memory capability, as shown in Figure~\ref{fig:framework}.





%


{In this case, we consider a heterogeneous local LoRA allocation, where the server assigns $c_i$ trainable LoRA layers for local training based on the client's capability.}
Specifically, we use an allocation map $m_i:=\{m_{i,(1)}, m_{i,(2)}, \cdots, m_{i,(l)}\}$ to represent the active status of the local LoRA layers for a client, where the element $m_{i,(j)} = 1$ if the $j$-th LoRA layer $\Delta\boldsymbol{\theta}_{i,(j)}$ is set to be locally trainable on client $i$ and $m_{i,(j)} = 0$ if the $j$-th LoRA layer is frozen, and $\sum(m_i)=c_i$. At each round $t$, with global parameters $\boldsymbol{\mathcal{\theta}}_g^{t}$, the local model of client $i$ is initialized as $\boldsymbol{\Phi}_i^{t, 0} =\{\boldsymbol{\mathcal{\theta}}_g^t, \boldsymbol{\Phi}^0\}$. Given the allocation map $m_i$, client $i$ freezes its $j$-th LoRA layer if $m_{i,(j)}=0$ and obtains its local trainable parameters, initialized as 
\begin{equation}\label{eq:local-init}
\boldsymbol{\mathcal{\theta}}_i^{t} := \{\boldsymbol{\theta}^t_{g,(j)} \mid j \in \{1, \cdots, l\} \text{ and } m_{i,(j)} = 1\}.
\end{equation}

After local training, each {selected} client uploads its local update 
$\boldsymbol{\delta}_i^{t+1} := \{\boldsymbol{\delta}_{i,(j)}^{t+1}\}_{j=1}^{l}$ 
to the server. These updates are then aggregated to update the global trainable parameters as,
\begin{equation}\label{eq:agg}
    \boldsymbol{\delta}_{g}^{t+1} = \left\{\frac{1}{\sum_{i\in\mathcal{S}^t} m_{i,(j)}} \sum_{i\in\mathcal{S}^t}m_{i,(j)} \boldsymbol{\delta}_{i,(j)}^{t+1} \right\}_{j=1}^{l}.
\end{equation}

\setlength{\textfloatsep}{3pt}
\begin{algorithm}[t]
\caption{Fed-HeLLo}\label{algorithm-ours}
\begin{algorithmic}[1]
\REQUIRE 
resource capability $c_i$, number of model layers $l$, heterogeneous layer allocation strategy HLA, number of communication rounds $T$.
\STATE Initial global model $\boldsymbol{\Phi}_g^0=\{\boldsymbol{\mathcal{\theta}}_g^0, \boldsymbol{\Phi}^0 \}$
\FOR{each clients $i \in [n]$ \textbf{in parallel}}
    \STATE Server broadcasts the initial global model $\boldsymbol{\Phi}_g^0$
    \STATE Client saves this model as their initial local models $\boldsymbol{\Phi}_i^0$
\ENDFOR
\FOR{$t=0$ to $T-1$}
    \STATE Server randomly samples a set of $s$ clients (denoted by $\mathcal{S}^t$), and obtain $c_i$
    \STATE Generate allocation maps $m_i^t= \text{HLA}(\boldsymbol{\theta}_{g}^t,\boldsymbol{\Phi}^0,c_i)$ for each client $i\in\mathcal{S}^t$ 
    \STATE Server broadcasts $\boldsymbol{\mathcal{\theta}}_g^t$ and $m_i^t$ to them
    \FOR{each clients $i \in \mathcal{S}^t$ \textbf{in parallel}}
        \STATE Initialize local trainable parameters $\boldsymbol{\mathcal{\theta}}_i^{t}$ as in Equation~\eqref{eq:local-init}
        \STATE $\boldsymbol{\delta}_i^{t}$ $\leftarrow \text{LocalUpdate}(\boldsymbol{\mathcal{\theta}}_i^{t}, \boldsymbol{\Phi}^0_i, D_i)$
    \ENDFOR
    \STATE $\boldsymbol{\delta}^{t+1}_g \leftarrow \text{Agg}(\boldsymbol{\theta}_{g}^t, \boldsymbol{\delta}^{t}, m^t)$ as in Equation~\eqref{eq:agg}
    \STATE $\boldsymbol{\mathcal{\theta}}_g^{t+1} = \boldsymbol{\mathcal{\theta}}_g^t + \boldsymbol{\delta}^{t+1}_g$
\ENDFOR
\STATE return $\boldsymbol{\Phi}_g^T=\{\boldsymbol{\mathcal{\theta}}_g^T, \boldsymbol{\Phi}^0 \}$
\end{algorithmic}
\end{algorithm}

Algorithm~\ref{algorithm-ours} outlines the overall framework of Fed-HeLLo. 
In each round, the server uses a HLA strategy to determine the allocation maps $m_i^t, i\in S^t$ for every selected client. In the following, we present the details of our HLA strategy. 
\subsection{Heterogeneous LoRA Allocation Methods}\label{sec:hla}

\begin{figure}[t]
\centering
\includegraphics[width=0.49\textwidth]{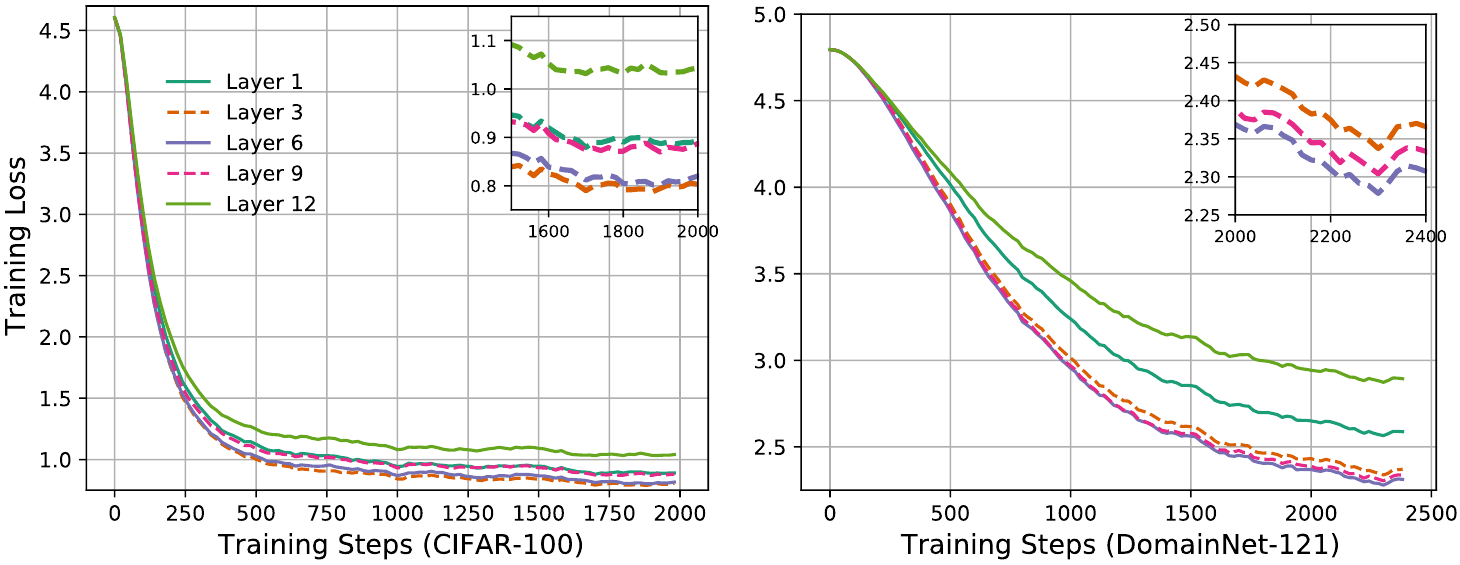}
\caption{{The convergence of fine-tuning different LoRA layers on ViT-base model in centralized settings across CIFAR100 and DomainNet-121 dataset.}}
\label{fig:layer_importance}
\end{figure}

{
Prior research~\cite{chen2022layer, ardakani2024slimfit,lodha2023surgical,pan2024lisa,gao2024higher,wang2020infinicache} has shown that different layers contribute unequally to model performance. As shown in Figure~\ref{fig:layer_importance}, training a different single LoRA layer results in different convergence speed, and this pattern varies for different tasks. 
%
In our method, we leverage the varying importance of LoRA layers to design HLA for clients with various memory capacities.}

\subsubsection{Fisher Information Matrix Score-based HLA for the Dynamic Layer Importance}\label{sec:FIM-HLA} 

To assess the importance of different layers, we adopt the Fisher Information Matrix (FIM) score~\cite{weiss1984application}.
%
Specifically, the FIM score measures how adjustments in model parameters affect the model's output. In practice, it is defined by calculating the average norm of the gradients,
\begin{equation}\label{eqn:FIM-definition}
\text{FIM}(\boldsymbol{\Phi}_l, \boldsymbol{\Phi}; D_\text{FIM}) = \frac{1}{|D_\text{FIM}|}\sum_{d\in D_\text{FIM}}\,\|\nabla_{\boldsymbol{\Phi}_l}\ell(\boldsymbol{\Phi}, d)\|^2_2\,,
\end{equation}
where $\boldsymbol{\Phi}$ denotes the model’s parameters, $l$ represents one layer of the model, $d$ denotes a data sample from a proxy dataset ${D_\text{FIM}}$, $\nabla\ell(\cdot)$ represents the model gradients, and $\|\cdot\|_2$ denotes the L2 norm.
Here, a higher FIM score indicates a more important layer. 

In our case, we calculate the FIM score on the server to assess the layer importance, and propose the FIM-HLA strategy.
%
%
Specifically, the server has a global model $\boldsymbol{\Phi}_g=\{\boldsymbol{\mathcal{\theta}},\boldsymbol{\Phi}^0\}$ and a proxy dataset $D_{\text{FIM}}$ sampled from server's test dataset without prior knowledge of the data distribution of clients~\cite{cao2020fltrust, park2021sageflow}.
The FIM score for each LoRA layer $j$ is obtained by $ \Gamma_{(j)}=\text{FIM}(\boldsymbol{\theta}_{(j)}, \boldsymbol{\Phi}_g, D_{\text{FIM}})$. Intuitively, client $i$ with resource capability $c_i$ should select the top $c_i$ important layers according to the FIM scores.

Instead of directly using the ranking of LoRA layers, 
we convert FIM scores into a smoother allocation probability distribution. 
%
Specifically, we divide clients' resource capacities $\{c_i\}_{i=1}^n$ into $k$ levels, $\{c_1, c_2, \dots, c_k\}$ in an ascending order. Each level $c_h, h\in[k]$ indicates the number of LoRA layers a client can train. 
If a client always pick the first $c_{h^{\prime}}$ layers to train, for the first $c_h$ LoRA layers, the corresponding allocation probability is,

\begin{equation}\label{eqn:g-star}
a_h=\frac{\sum_{h^{\prime}=1}^{k}{\mathbf{1}(c_{h^{\prime}} \geq c_h)}}{\sum_{h^{\prime}=1}^{k}{c_{h^{\prime}} r_{h^{\prime}}}},h\in[k],
\end{equation}
where $r_h$ represents the ratio of clients which has the capability of $c_h$. This gives us a base probability $\{a_1, a_2, \dots, a_k\}$ for different resource levels. 
%
%
Next, we cluster~\cite{hartigan1979algorithm} the FIM scores $\{\Gamma_{(j)}\}_{j=1}^l$ of LoRA layers into $k$ groups $\{g_1, g_2, \dots g_k\}$ in a
descending order. For the layer with a higher FIM score, it should has a larger allocation probability. Therefore, 
for each layer $j \in g_h, h\in [k]$, its allocation probability is set by $a_{j}$. The allocation probabilities across all layers are then normalized so that their sum equals one.
%
%
Finally, the allocation map $m_i$ for client $i$ can be determined individually by randomly sampling $c_i$ LoRA layers without replacement based on the allocation probability distribution.

\subsubsection{Randomized Geometrically-Defined Heterogeneous LoRA Allocation for the Intrinsic Layer Importance}\label{sec:GD-DPP-HLA}

\begin{figure}[]
\centering
\includegraphics[width=0.38\textwidth]{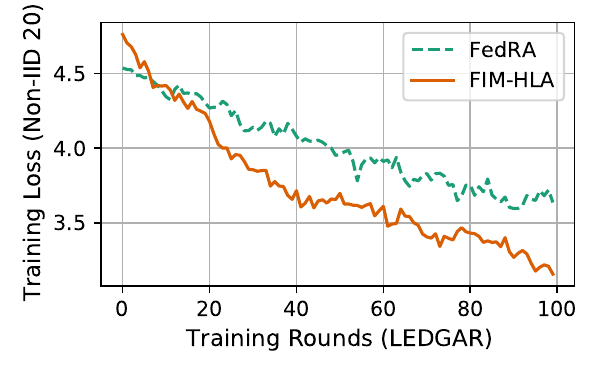}
\caption{{The convergence of FedRA and FIM-HLA on LEDGAR dataset.}}
\label{fig:fim-drawback}
\end{figure}

{
In Fed-HeLLo, FIM-HLA is conducted every round to generate allocation maps for clients. However, during the early training rounds, FIM scores do not provide an accurate estimation about the training state of the global model.
To explore this, we conduct an experiment on LEDGAR dataset, comparing the convergence speed of our proposed FIM-HLA with a random allocation baseline, FedRA~\cite{su2023fedra}. As shown in Figure~\ref{fig:fim-drawback}, FIM-HLA achieves faster overall convergence compared to FedRA, but FedRA initially converges faster during the first 10 rounds. This discrepancy may result from FIM-HLA's difficulty in capturing meaningful layer importance information early in training. To address this issue, we introduce a HLA method that provides an effective initial LoRA allocation by considering intrinsic layer importance.}

{
The intrinsic layer importance represents the natural role of each layer in feature extraction~\cite{pan2024lisa,gao2024higher}. For example, different tasks and datasets often require models to focus on specific feature hierarchies, where shallow layers capture edges, textures, and spatial structures, deep layers extract semantic meaning, object relationships, and contextual understanding. Object-centric datasets such as CIFAR-100~\cite{krizhevsky2009learning} and ImageNet~\cite{deng2009imagenet} emphasize deep-layer semantic features, whereas datasets like the Describable Textures Dataset (DTD)~\cite{cimpoi2014describing} prioritize shallow-layer textural details.} 
%
Inspired by these, we propose a Geometrically-Defined HLA (GD-HLA) strategy. This strategy allocates trainable LoRA layers to clients according to specific {patterns} that are effective at the global level, where the patterns are designed to consider the intrinsic LoRA layers importance. Specifically, we define four GD-HLAs, each resulting in a distinct global allocation pattern: \textit{Triangle} ($\rhd$), \textit{Inverted Triangle} ($\lhd$), \textit{Bottleneck} ($\bowtie$), and \textit{Uniform} ($\approx$). In the \textit{Triangle} pattern ($\rhd$), clients prioritize training on shallower LoRA layers, allocating more resources to these layers due to their significant impact on coarse-grained feature modeling. In contrast, the \textit{Inverted Triangle} pattern ($\lhd$) focuses on deeper LoRA layers, emphasizing the development of the ability to handle more semantic features. The \textit{Bottleneck} pattern ($\bowtie$) represents a trade-off between these two allocations, acknowledging the significance of both shallow and deep LoRA layers in feature modeling~\cite{zeiler2014visualizing}. Finally, the \textit{Uniform} pattern ($\approx$) assigns uniform significance to features across all layers, resulting in a uniform allocation by randomly selecting $c_i$ layers. 

Each strategy corresponds to a specific allocation mask for clients $i$, formalized as follows:

\begin{align}\label{eq:gd-HLA-assign}
m_i^\rhd    & = \left\{\underbrace{1,\cdots,1}_{c_i},\underbrace{0,\cdots,0}_{l-c_i}\right\}, \nonumber\\
m_i^\lhd    & = \left\{\underbrace{0,\cdots,0}_{l-c_i},\underbrace{1,\cdots,1}_{c_i}\right\}, \\
m_i^{\bowtie} & = \left\{\underbrace{1,\cdots,1}_{\left \lceil {c_i}/{2}\right \rceil },\underbrace{0,\cdots,0}_{l-c_i},\underbrace{1,\cdots,1}_{\left \lfloor {c_i}/{2}\right \rfloor}\right\}, \nonumber\\
m_i^{\approx} & = \left\{0,1\right\}^{l} \sim \text{Uniform}\left(c_i/l\right).\nonumber
\end{align}

Here, we extend the fixed GD-HLA to a randomized version, named Randomized Geometrically-Defined HLA (RGD-HLA).
Specifically, using the allocation maps $\{m_i^\star\}_{i=1}^n$ generated by a pre-defined GD-HLA strategy $\star$, a prior allocation probability for each layer $j$ is calculated as 
\begin{equation}\label{eqn:dpp-HLA1}
g_{(j)}^{\star} = \frac{\sum_{i=1}^{n} m_{i,(j)}^{\star}}{\sum_{j=1}^l \sum_{i=1}^{n} {m_{i, (j)}^{\star}}},
\end{equation}
where $\sum_{j=1}^l g_{(j)}^{\star} =1$. 
{
Based on the allocation probability distribution $\{g_1, g_2, \dots, g_l^{\star}\}$, the server conducts a random sampling by selecting $c_i$ LoRA layers without replacement to obtain the allocation map for each client $i$.} This approach introduces randomness into the allocation of trainable LoRA layers based on a geometrically-defined prior probability, allowing resource-constrained clients the chance to train different layers, which can potentially improve fine-tuning performance. Note that despite the randomness at the individual client level, the overall distribution of layer allocations across all clients adheres to a stable and predictable pattern predefined by $\star$.


\begin{table*}[t]
  \centering
  \caption{{Main results of Fed-HeLLo compared with baselines on CIFAR-100 dataset.}}
    \scalebox{1.1}{
    \begin{tabular}{c|c|cccc|cccc}
    \toprule[1pt]
    \multicolumn{2}{c|}{\multirow{2}[3]{*}{\textbf{Methods}}} & \multicolumn{4}{c|}{\textbf{Model Performance (Accuracy)}} & \multicolumn{1}{c}{\multirow{2}[0]{*}{\shortstack{\textbf{Avg Back. Comp.}\\\textbf{Cost (TFLOPs)}}}} & \multicolumn{1}{c}{\multirow{2}[0]{*}{\shortstack{\textbf{Avg Memory}\\\textbf{Cost (GB)}}}} & \multicolumn{1}{c}{\multirow{2}[0]{*}{\shortstack{\textbf{Avg Comm.}\\\textbf{Cost (MB/R)}}}} & \multicolumn{1}{c}{\multirow{2}[0]{*}{\shortstack{\textbf{Avg Training}\\\textbf{Time (Sec/R)}}}}  \\
\cmidrule{3-6}    \multicolumn{2}{c|}{} & \textbf{IID} & \textbf{20/1.0} & \textbf{10/1.0} & \textbf{Avg} &       &       &       &  \\
    \midrule
    Straggler & - & 68.15 & 60.48 & 55.88 & 61.50 &1.62       & 10.52 & 3.53      &6.38  \\ \midrule
    Exclusive & \multicolumn{1}{c|}{\multirow{6}[0]{*}{\shortstack{RoC\\6:3:1}}} & 78.33 & 4.49  & 7.13  & 29.98 &  3.25     & 12.29 & 4.71      &12.99\\
    HETLoRA &       & 80.10 & 64.36 & 53.20 & 65.88 & 3.22      & 12.25 & 2.85      &9.07  \\
    FlexLoRA &       & 44.77 & 44.36 & 38.40 & 42.51 & 3.22      & 12.25 & 57.12      &14.91  \\
    FLoRA &       & 3.56  & 2.39  & 2.88  & 2.94  & 3.22      & 12.25 & 5.47      &11.62  \\
    FedRA &       & 86.19 & 81.08 & 74.77 & 80.68 & 2.03      & 10.47 & 3.83      &10.17  \\
    \textbf{Fed-HeLLo} &       & \cellcolor[rgb]{ .749,  .749,  .749}86.50 & \cellcolor[rgb]{ .749,  .749,  .749}82.57 & \cellcolor[rgb]{ .749,  .749,  .749}78.49 & \cellcolor[rgb]{ .749,  .749,  .749}82.52 & 2.03      & 10.74 &3.83       &9.83  \\ \midrule
    Exclusive & \multicolumn{1}{c|}{\multirow{7}[0]{*}{\shortstack{RoC\\1:1:1}}}  & 85.20 & 81.14 & 70.26 & 78.86 & 3.25      & 12.19 & 4.71      & 11.87 \\
    HETLoRA &       & 85.56 & 81.06 & 69.69 & 78.77 & 3.23      & 12.26 & 3.41      &7.83  \\
    FlexLoRA &       & 42.35 & 40.47 & 36.32 & 39.71 & 3.23      & 12.26 & 57.67      &21.23  \\
    FLoRA &       & 4.04  & 3.58  & 2.29  & 3.30 & 3.23      & 12.26 & 11.53      & 12.92 \\
    FedRA &       & 87.06 & 84.05 & 78.51 & 83.20 & 2.45      & 11.08 &4.14       &9.52  \\
    \textbf{Fed-HeLLo} &       & \cellcolor[rgb]{ .749,  .749,  .749}87.11 & \cellcolor[rgb]{ .749,  .749,  .749}84.64 & \cellcolor[rgb]{ .749,  .749,  .749}79.76 & \cellcolor[rgb]{ .749,  .749,  .749}83.84 &2.45       & 11.19 &4.14       & 5.01 \\ \bottomrule[1pt]
    \end{tabular}%
    }
  \label{tab:main-cifar}%
  \arrayrulecolor{black}
\end{table*}%

\begin{table*}[t]
  \centering
  \caption{{Main results of Fed-HeLLo compared with baselines on LEDGAR dataset.}}
  \scalebox{1.1}{
    \begin{tabular}{c|c|cccc|cccc}
      \toprule[1pt]
    \multicolumn{2}{c|}{\multirow{2}[3]{*}{\textbf{Methods}}} & \multicolumn{4}{c|}{\textbf{Model Performance (Macro-F1)}} & \multicolumn{1}{c}{\multirow{2}[0]{*}{\shortstack{\textbf{Avg Back. Comp.}\\\textbf{Cost (TFLOPs)}}}} & \multicolumn{1}{c}{\multirow{2}[0]{*}{\shortstack{\textbf{Avg Memory}\\\textbf{Cost (GB)}}}} & \multicolumn{1}{c}{\multirow{2}[0]{*}{\shortstack{\textbf{Avg Comm.}\\\textbf{Cost (MB/R)}}}} & \multicolumn{1}{c}{\multirow{2}[0]{*}{\shortstack{\textbf{Avg Training}\\\textbf{Time (Sec/R)}}}}  \\
\cmidrule{3-6}    \multicolumn{2}{c|}{} & \textbf{IID} & \textbf{20/1.0} & \textbf{10/1.0} & \textbf{Avg} &       &       &       &  \\ \midrule
    Straggler & - & 56.82 & 51.05 & 44.05 & 50.64 & 1.62      &10.61       & 3.53      &6.58  \\ \midrule
    Exclusive & \multicolumn{1}{c|}{\multirow{6}[0]{*}{\shortstack{RoC\\6:3:1}}} & 61.25 & 44.72 & 28.48 & 44.81 & 3.25      & 12.38      & 4.71      &12.99  \\
    HETLoRA &       & 53.00 & 39.09 & 26.90 & 39.66 & 3.22      &12.35       & 2.87      &9.07  \\
    FlexLoRA &       & 29.08 & 4.08  & 5.01  & 12.72 &3.22       &12.35       & 57.13      &14.91  \\
    FLoRA &       & 0.06  & 0.04  & 0.08  & 0.06  & 3.22      &12.35       & 5.65      &11.62  \\
    FedRA &       & 66.74 & 61.73 & 55.71 & 61.39 & 2.04      &10.57       & 3.83      &10.17  \\
    \textbf{Fed-HeLLo} &       & \cellcolor[rgb]{ .749,  .749,  .749}67.00 & \cellcolor[rgb]{ .749,  .749,  .749}65.58 & \cellcolor[rgb]{ .749,  .749,  .749}61.04 & \cellcolor[rgb]{ .749,  .749,  .749}64.54 &  2.04     & 10.74      & 3.83      & 9.83 \\ \midrule
    Exclusive & \multicolumn{1}{c|}{\multirow{6}[0]{*}{\shortstack{RoC\\1:1:1}}} & 66.17 & 57.41 & 13.20 & 45.59 & 3.25      &12.38       & 4.71      & 8.45 \\
    HETLoRA &       & 62.88 & 54.50 & 45.78 & 54.38 &3.23       & 12.36      & 3.40      & 7.09 \\
    FlexLoRA &       & 30.95 & 1.19  & 1.83  & 11.32 &3.23       &12.36       & 57.66      & 13.33 \\
    FLoRA &       & 0.08  & 0.04  & 0.01  & 0.04  & 3.23      & 12.36      & 11.53      &12.18  \\
    FedRA &       & 66.63 & 62.61 & 56.39 & 61.87 & 2.45      & 11.16      & 4.13      & 6.54 \\
    \textbf{Fed-HeLLo} &       & \cellcolor[rgb]{ .749,  .749,  .749}66.75 & \cellcolor[rgb]{ .749,  .749,  .749}67.10 & \cellcolor[rgb]{ .749,  .749,  .749}64.27 & \cellcolor[rgb]{ .749,  .749,  .749}66.04 & 2.45      &11.28       & 4.13      & 5.75 \\
    \bottomrule[1pt]
  \end{tabular}%
    }
  \label{tab:main-ledgar}%
  \arrayrulecolor{black}
  
\end{table*}%

\subsubsection{Co-design of FIM-HLA and RGD-HLA}\label{sec:design-framework}

To consider both the dynamic and intrinsic importance of different LoRA layers, we co-design FIM-HLA and RGD-HLA to finalize our HLA strategy. 
%
%
{
Specifically, our framework conducts RGD-HLA during the first $T_{\text{RGD}}$ rounds, which helps the model training to ``warm-start''.}
Then, at every $T_{\text{FIM}}$ rounds, FIM-HLA is used to update the allocation probability distribution for each LoRA layer. 

\subsection{Computational Complexity and Communicational Cost}\label{sec:complexity}

{\textbf{Computational Complexity.}} 
{During the FL process, the computational complexity can be analyzed in terms of local training, model aggregation, and other operations~\cite{wu2022node,zhou2022exact}. Here, we compare our method with FedAvg~\cite{mcmahan2017communication}.}

{During local training, the computational complexity is determined by forward and backward computations. For FedAvg, these operations have a computational complexity of $O(\tau l (d+R)^2 N s)$, where $\tau$ represents the number of local epochs per round, $l$ represents the number of layers, $d$ represents the number of parameters for pre-trained model in each layer, $R$ denotes the number of parameters for the LoRA layer, $N$ represents the number of training samples, and $s$ is the number of clients. The term $(d+R)^2$ represents the number of operations needed for matrix multiplications. 
For Fed-HeLLo, only $c$ LoRA layers are trainable, which results in a computational complexity of $O(\frac{\tau c (d+R)^2 N s}{l})$}.

{During model aggregation, the summing and averaging operations for Fed-HeLLo accounts for a computational complexity of $O(s l R)$, which is the same for FedAvg.}

{For Fed-HeLLo, there is an extra FIM score calculation cost every $T_{\text{FIM}}$ rounds on the server. The computational complexity is $O(\frac{l (d+R)^2 N_{\text{FIM}}T_{\text{FIM}}}{T} + \frac{R^2T_{\text{FIM}}}{T})$, where $N_{\text{FIM}}$ represents the number of samples used for calculating the FIM score, and $T$ represents the total training round. The first term is the computational complexity for obtaining the gradients. The second term represents the computational complexity of calculating the gradient norms.}

{Overall, the computational complexity of FedAvg is given by $O(\tau l (d+R)^2 N s + s l R)$. For Fed-HeLLo, it is $O(\frac{\tau c (d+R)^2 N s}{l} + s l R + \frac{l (d+R)^2 N_{\text{FIM}}T_{\text{FIM}}}{T} + \frac{R^2T_{\text{FIM}}}{T})$. Since $\frac{T_{\text{FIM}}}{T}$ is small, the last two terms can be neglected. Additionally, as aggregation operations are less computationally expensive than training, the second term is also omitted. As a result, Fed-HeLLo achieves reduced computational complexity by training fewer layers.}

{\textbf{Communication Cost}}
%
%
{In FedAvg, the full trainable LoRA layers are transmitted from clients to the server and then broadcast to all selected clients, resulting in a communication cost per round of $O(2 l R s)$. In Fed-HeLLo, allocation maps are transmitted every $T_{\text{FIM}}$ rounds, leading to a total communication cost of $O(2 c R s + \frac{l s T_{\text{FIM}}}{T})$. Since the communication cost for FIM scores is negligible, reducing the number of trainable LoRA layers decreases the communication cost, demonstrating the efficiency and practicality of our method in FL scenarios.}

\section{Experiments} 
In this section, we evaluate and compare the performance of our proposed method, Fed-HeLLo, with other baseline methods across various datasets and data distribution settings.

\subsection{Experimental Settings}\label{sec:exp}
We conduct experiments on both visual and language modalities. For visual tasks, we use CIFAR-100 dataset~\cite{krizhevsky2009learning} and a crafted version of DomainNet dataset~\cite{peng2019moment}, referred to as DomainNet-121. {For language tasks, we use LEDGAR dataset from LexGLUE~\cite{chalkidis-etal-2021-lexglue} for paragraph classification, as well as two instruction fine-tuning datasets: Natural Instruction~\cite{wang2022super} and Dolly-15K~\cite{ouyang2022training}.} We simulate three levels of data heterogeneity ranging from IID to extreme Non-IID by varying the data quantities, labels, or domains of clients. For example, a ``20/1.0'' Non-IID setting means that each client has samples from 20 categories, with quantities skewed by a Dirichlet distribution with $\alpha=1.0$; a ``25/D1'' Non-IID setting indicates that each client has 25 categories and one domain per category. 

For visual tasks, we employ ViT-base~\cite{dosovitskiy2020image}, a 12-layer transformer encoder model designed for image processing. {For language tasks, we conduct experiments on three models. For LEDGAR dataset, we utilize BERT-base~\cite{devlin2019bert}, a 12-layer transformer model pre-trained on a comprehensive English corpus in a self-supervised manner. For the Natural Instruction and Dolly-15K dataset, we use the Llama-like model DataJuicer-1B~\cite{chen2024data} and OPT-1.3B~\cite{zhang2022opt} as the pre-trained models, respectively. For all these models, LoRA layers are applied to both the ``query'' and ``value'' modules in each transformer layer. The LoRA rank is set to 16 for ViT-base and BERT-base, and 8 for DataJuicer-1B and OPT-1.3B. The scaling parameter and dropout probability are set to 16 and 0.1, respectively.}

{For evaluation, we use top-1 accuracy for CIFAR-100 and DomainNet-121, Macro F1 score for LEDGAR~\cite{chalkidis-etal-2021-lexglue}, and Rouge-L score~\cite{lin2004rouge} for Natural Instruction and Dolly-15K. The batch sizes for local training on CIFAR100, LEDGAR, DomainNet-121, Natural Instruction, and Dolly-15K are 128, 128, 32, 1, and 1, respectively.}

\begin{table*}[t]
  \centering
  \caption{{Main results of Fed-HeLLo compared with baselines on Natural Instruction dataset.}}
  \scalebox{1.1}{
    \begin{tabular}{c|c|cccc|cccc}
      \toprule[1pt]
    \multicolumn{2}{c|}{\multirow{2}[3]{*}{\textbf{Methods}}} & \multicolumn{4}{c|}{\textbf{Model Performance (Rouge-L)}} & \multicolumn{1}{c}{\multirow{2}[0]{*}{\shortstack{\textbf{Avg Back. Comp.}\\\textbf{Cost (TFLOPs)}}}} & \multicolumn{1}{c}{\multirow{2}[0]{*}{\shortstack{\textbf{Avg Memory}\\\textbf{Cost (GB)}}}} & \multicolumn{1}{c}{\multirow{2}[0]{*}{\shortstack{\textbf{Avg Comm.}\\\textbf{Cost (MB/R)}}}} & \multicolumn{1}{c}{\multirow{2}[0]{*}{\shortstack{\textbf{Avg Training}\\\textbf{Time (Sec/R)}}}}  \\
\cmidrule{3-6}    \multicolumn{2}{c|}{} & \textbf{IID} & \textbf{20/1.0} & \textbf{15/1.0} & \textbf{Avg} &       &       &       &  \\ \midrule
    Straggler & - &59.71	&59.37	&59.11	&59.39  & 1.83      & 11.16      &  9.43     & 99.07 \\ \midrule
    Exclusive & \multicolumn{1}{c|}{\multirow{6}[0]{*}{\shortstack{RoC\\6:3:1}}} &59.58	&59.34	&59.42	&59.44 & 3.67      &12.40       & 12.58      &200.20  \\
    HETLoRA &       &60.65	&60.00	&60.21	&60.28 & 3.67      & 12.37      & 7.43      &153.40  \\
    FlexLoRA &      &61.93	&60.10	&\cellcolor[rgb]{ .749,  .749,  .749}60.86	&60.96  & 3.67      &12.37       & 806.44      &205.40  \\
    FLoRA &       &48.15	&48.69	&47.84	&48.22  & 3.67      & 12.37      & 7.36      &165.11  \\
    FedRA &      &60.91	&60.84	&60.50	&60.75  & 2.26      & 11.25      & 10.16      &101.46  \\
    \textbf{Fed-HeLLo} &       & \cellcolor[rgb]{ .749,  .749,  .749}66.18 & \cellcolor[rgb]{ .749,  .749,  .749}61.33 & 59.94 & \cellcolor[rgb]{ .749,  .749,  .749}62.48 & 2.26      & 11.28      & 10.16      &100.59  \\ \midrule
    Exclusive & \multicolumn{1}{c|}{\multirow{6}[0]{*}{\shortstack{RoC\\1:1:1}}} &63.03	&59.96	&59.58	&60.85 & 3.67      &12.40       & 12.58      &160.50  \\
    HETLoRA &       & 62.47	&60.39	&60.24	&61.03 &  3.67     & 12.38      & 8.92      &159.20  \\
    FlexLoRA &     & 61.63	&60.67	&60.49	&60.93  & 3.67      & 12.38      & 807.94      &191.79  \\
    FLoRA &       &48.05	&46.38	&48.88	&47.77 & 3.67      & 12.38      &  16.82     &143.36  \\
    FedRA &     & 61.84	  &61.41	&60.49	&61.24  & 2.76      &11.64       & 11.02      &98.55  \\
    \textbf{Fed-HeLLo} &       & \cellcolor[rgb]{ .749,  .749,  .749}68.22 & \cellcolor[rgb]{ .749,  .749,  .749}62.03 & \cellcolor[rgb]{ .749,  .749,  .749}60.89 & \cellcolor[rgb]{ .749,  .749,  .749}63.71 & 2.76      & 11.64      & 11.02      & 109.79 \\
    \bottomrule[1pt]
  \end{tabular}%
    }
  \label{tab:main-natural_instruction}%
  \arrayrulecolor{black}
  
\end{table*}%

\begin{table*}[t]

  \centering
  \caption{{Main results of Fed-HeLLo compared with baselines on Dolly-15K dataset.}}
  \scalebox{1.1}{
    \begin{tabular}{c|c|cccc|cccc}
      \toprule[1pt]
    \multicolumn{2}{c|}{\multirow{2}[3]{*}{\textbf{Methods}}} & \multicolumn{4}{c|}{\textbf{Model Performance (Rouge-L)}} & \multicolumn{1}{c}{\multirow{2}[0]{*}{\shortstack{\textbf{Avg Back. Comp.}\\\textbf{Cost (TFLOPs)}}}} & \multicolumn{1}{c}{\multirow{2}[0]{*}{\shortstack{\textbf{Avg Memory}\\\textbf{Cost (GB)}}}} & \multicolumn{1}{c}{\multirow{2}[0]{*}{\shortstack{\textbf{Avg Comm.}\\\textbf{Cost (MB/R)}}}} & \multicolumn{1}{c}{\multirow{2}[0]{*}{\shortstack{\textbf{Avg Training}\\\textbf{Time (Sec/R)}}}}  \\
\cmidrule{3-6}    \multicolumn{2}{c|}{} & \textbf{IID} & \textbf{3/1.0} & \textbf{1/1.0} & \textbf{Avg} &       &       &       &  \\ \midrule
    Straggler & - &57.89	&57.76	&57.83	&57.82   &  0.86     & 10.49      & 9.43      & 51.69 \\ \midrule
    Exclusive & \multicolumn{1}{c|}{\multirow{6}[0]{*}{\shortstack{RoC\\6:3:1}}} &57.98	&58.80	&57.92	&58.23  &  1.73     & 11.72      & 12.58      &108.41  \\
    HETLoRA &    &58.44	&58.92	&53.03	&56.79   & 1.73      & 11.70      &7.60       & 86.48 \\
    FlexLoRA &   &58.86	&58.84	&59.14	&58.94     & 1.73      & 11.70      & 806.62      &138.89  \\
    FLoRA &      &40.69	&41.47	&44.68	&42.28   & 1.73      & 11.70      & 7.58      & 114.42 \\
    FedRA &     &59.32	&59.28	&58.76	&59.12   & 1.06      & 10.57      & 10.14      &103.46  \\
    \textbf{Fed-HeLLo} &       & \cellcolor[rgb]{ .749,  .749,  .749}59.80 & \cellcolor[rgb]{ .749,  .749,  .749}59.90 & \cellcolor[rgb]{ .749,  .749,  .749}59.45 & \cellcolor[rgb]{ .749,  .749,  .749}59.71 &  1.06     & 10.61      & 10.14      &94.25  \\ \midrule
    Exclusive & \multicolumn{1}{c|}{\multirow{6}[0]{*}{\shortstack{RoC\\1:1:1}}} &59.38	&59.37	&58.58	&59.11  & 1.73      & 11.72      &  12.58     &131.39  \\
    HETLoRA &     & 59.36	&59.34	&59.08	&59.26  &  1.73     &11.71       & 9.45      &90.14  \\
    FlexLoRA &    &58.82	&58.98	&58.68	&58.82   & 1.73      & 11.71      & 808.47      &133.51  \\
    FLoRA &       &42.19	&44.72	&44.70	&43.87 &  1.73     &11.71       & 19.03      &112.37  \\
    FedRA &       &59.38	&59.38	&59.07	&59.27  &  1.39     &11.11       & 11.33      &104.44  \\
    \textbf{Fed-HeLLo} &       & \cellcolor[rgb]{ .749,  .749,  .749}59.41 & \cellcolor[rgb]{ .749,  .749,  .749}59.40 & \cellcolor[rgb]{ .749,  .749,  .749}59.20 & \cellcolor[rgb]{ .749,  .749,  .749}59.33 & 1.39      & 11.14      &11.33       & 104.48 \\
    \bottomrule[1pt]
  \end{tabular}%
    }
  \label{tab:main-dolly15k}%
  \arrayrulecolor{black}
\end{table*}%

{For the FL settings, we consider $n=100$ clients for classification tasks (CIFAR-100, LEDGAR, and DomainNet-121) and $n=50$ for instruction fine-tuning tasks (Natural Instruction and Dolly-15k) unless otherwise specified. The local epoch $\tau$ and the number of training rounds $T$ are set to 1 and 500 for classification tasks, and 1 and 25 for instruction fine-tuning tasks, respectively. At each round, the server randomly selects $s=10$ clients to perform local fine-tuning, using the AdamW optimizer as the local solver.}

{We consider two settings for the number of clients. In a practical scenario with more low-resource devices, the ratio of clients is set to 6:3:1, with each group capable of training $\frac{1}{2}$, $\frac{3}{4}$, or all LoRA layers, corresponding to three levels of resource capabilities. We also consider an evenly distributed case (1:1:1 ratio of clients), where clients are divided into three resource levels, each capable of training $\frac{1}{2}$, $\frac{3}{4}$, or all LoRA layers. 
For the proxy dataset used in FIM-HLA, we randomly sample 100, 50, 1000, 50, and 50 samples from the test dataset of CIFAR100, LEDGAR, DomainNet-121, Natural Instruction, and Dolly-15K datasets, respectively, and exclude these samples throughout the entire process to ensure fairness.
In classification experiments, we set both $T_{\text{RGD}}$ and $T_{\text{FIM}}$ to 50 for Fed-HeLLo, and $T_{\text{RGD}}$=0 and $T_{\text{FIM}}$=1 for FIM-HLA. In instruction fine-tuning experiments, we set  $T_{\text{RGD}}$=1 and $T_{\text{FIM}}$=2 for Fed-HeLLo, and same as before for FIM-HLA.}

\subsection{Baselines}
To demonstrate the effectiveness of Fed-HeLLo, we compare it with several baselines as follows:
\textbf{Straggler Learning} allocates $\min\{c_i\}_{i=1}^n$ trainable layers for all clients based on the straggler's resource capability, i.e., in our case $c_i=\frac{l}{2}$, $\forall i\in[n]$.
\textbf{Exclusive Learning} excludes clients that are unable to train all LoRA layers.
\textbf{HETLoRA~\cite{cho2023heterogeneous}} is a federated LoRA-based fine-tuning framework that supports heterogeneous LoRA ranks across clients. 
{For HETLoRA, we align the number of clients across three levels of resource capabilities with our method's settings, with each group of clients capable of training LoRA layers at ranks 1, 4, or 16 for classification tasks and 1, 1, or 8 for instruction fine-tuning tasks.
{\textbf{FLoRA}}~\cite{wang2024flora} assigns lower ranks to resource-constrainted clients and shares stacked LoRA modules for updates.
{\textbf{FlexLoRA}}~\cite{bai2024flexlora} improves aggregation in rank-based methods using SVD.
}
\textbf{FedRA~\cite{su2023fedra}} is a federated LoRA-based fine-tuning framework that generates a random allocation map to allocate trainable LoRA layers to clients for each round. 

\begin{table}[t]
  \centering
  \caption{{Main results of Fed-HeLLo compared with baselines on DomainNet-121 dataset.}}
  \scalebox{1.1}{
    \begin{tabular}{c|c|cccc}
    \toprule[1pt]
    \multicolumn{2}{c|}{\multirow{2}[2]{*}{\textbf{Methods}}} & \multicolumn{4}{c}{\textbf{Model Performance (Accuracy)}} \\
    \cmidrule{3-6}    \multicolumn{2}{c|}{} & \textbf{IID} & \textbf{25/D1} & \textbf{15/D1} & \textbf{Avg} \\ \midrule
    Straggler & \multicolumn{1}{c|}{-} & 49.03 & 34.87 & 29.20 & 37.70     \\
    \midrule
    Exclusive & \multicolumn{1}{c|}{\multirow{6}[0]{*}{\shortstack{RoC\\6:3:1}}} & 54.04 & 29.69 & 21.43 & 35.05    \\
    HETLoRA &       & 54.43 & 27.98 & 13.52 & 31.97      \\
    {FlexLoRA} &      &{31.12}  &{3.41}  &{0.99}  &{11.84}     \\
    {FLoRA}    &      &{1.51}  &{1.15}  &{1.01}  &{1.22}   \\
    FedRA &       & 65.39 & 52.39 & 48.26 & 55.34  \\
    \textbf{Fed-HeLLo} &       & \cellcolor[rgb]{ .749,  .749,  .749}67.25 & \cellcolor[rgb]{ .749,  .749,  .749}54.44 & \cellcolor[rgb]{ .749,  .749,  .749}50.22 & \cellcolor[rgb]{ .749,  .749,  .749}57.30       \\
    \midrule
    Exclusive & \multicolumn{1}{c|}{\multirow{6}[0]{*}{\shortstack{RoC\\1:1:1}}} & 65.23 & 51.70 & 46.58 & 54.50       \\
    HETLoRA &       & 64.26 & 48.92 & 35.34 & 49.50      \\
    {FlexLoRA} &      &{23.05}  &{1.21}  &{1.21}  &{8.44}   \\
    {FLoRA} &         &{1.79}  &{1.01}  &{1.07}  &{1.29}   \\
    FedRA &       & 68.53 & 55.64 & 51.07 & 58.41  \\
    \textbf{Fed-HeLLo} &       & \cellcolor[rgb]{ .749,  .749,  .749}68.98 & \cellcolor[rgb]{ .749,  .749,  .749}58.01 & \cellcolor[rgb]{ .749,  .749,  .749}53.74 & \cellcolor[rgb]{ .749,  .749,  .749}60.24       \\
    \bottomrule[1pt]
    \end{tabular}%
    }
  \label{tab:main-domainnet121}%
  \arrayrulecolor{black}
  
\end{table}%

\subsection{Experimental Results}\label{exp:main-results}
We evaluate the effectiveness of our Fed-HeLLo framework on five datasets under three levels of data heterogeneity, comparing it with baseline methods. The main experimental results are summarized in Tables~\ref{tab:main-cifar}, \ref{tab:main-ledgar}, \ref{tab:main-natural_instruction}, \ref{tab:main-dolly15k} and \ref{tab:main-domainnet121}, where the best results among all methods are highlighted in \colorbox[rgb]{ .749,  .749,  .749}{grey}, and the average performance across all data settings is also reported. 
{Additionally, we provide a comparison of several resource indicators, including the average backward computational cost per step across all clients, average memory usage per step, average communication cost per round, and average wall-clock training time per round.
RoC represents the ratio of heterogeneous resource clients.}


\textbf{Quantitative Results on CIFAR-100:} We compare our method Fed-HeLLo with two baseline methods (i.e., Straggler and Exclusive Learning) and four SOTA heterogeneous federated LoRA-based fine-tuning methods (i.e., HETLoRA, FlexLoRA, FLoRA, and FedRA). Table~\ref{tab:main-cifar} reports the comparison results. In CIFAR-100 experiments, Fed-HeLLo demonstrates substantial improvements across various settings. Under the 6:3:1 RoC distribution, Fed-HeLLo achieves an average accuracy of 82.52\%, surpassing Straggler Learning by +21.02\% and Exclusive Learning by +52.54\%. {Compared to the rank-based method HETLoRA, the performance gains are even more significant (+6.40\%, +18.21\%, +25.29\%) as the data heterogeneity becomes more extreme. FlexLoRA and FLoRA result in lower overall performance, indicating that their SVD and stacking techniques hinder the global model's convergence. This is likely due to their noise-alleviation aggregation being ineffective when not all LoRA layers are trainable.} Additionally, our method achieves substantial improvements over the SOTA method FedRA, demonstrating the effectiveness of leveraging both dynamic and intrinsic layer importance in our framework. Similarly, under the 1:1:1 RoC setting, our method continues to outperform other methods, achieving +1.25\% and +10.07\% improvements over FedRA and HETLoRA, respectively, in the ``10/1.0'' Non-IID setting. Notably, Exclusive Learning outperforms HETLoRA in this setting, indicating that rank-based method can be detrimental during training. These results underscore the robustness of Fed-HeLLo in effectively managing resource heterogeneity in CIFAR-100 dataset. {Regarding resource efficiency, under the 6:3:1 RoC distribution, rank-based methods such as HETLoRA, FlexLoRA, and FLoRA achieve minimal computational and memory savings compared to full LoRA layer fine-tuning. In contrast, our method reduces computational cost by 38.41\% and saves an average of 1.55 GB of memory. Additionally, our approach requires only 3.83 MB of communication per round, significantly lower than FlexLoRA, which needs to transmit full-rank matrices for each client.
}

\textbf{Quantitative Results on LEDGAR:} On LEDGAR dataset~\ref{tab:main-ledgar}, our method consistently demonstrates strong performance. Under the 6:3:1 RoC distribution, Fed-HeLLo improves the macro F1 score by +13.90\% compared to Straggler Learning and +19.73\% compared to Exclusive Learning. Notably, in the most extreme Non-IID scenario (10/1.0), Fed-HeLLo continues to outperform other methods, with a +5.33\% improvement over FedRA and a +34.14\% improvement over HETLoRA. {FlexLoRA and FLoRA also perform poorly in this scenario. It also indicates that these rank-based methods are not scalable to mid-sized FMs due to the limited prior information in the pre-trained weights.}
In the 1:1:1 RoC scenario, our method maintains all macro F1 scores above 60\%, and outperforms the SOTA methods HETLoRA and FedRA by +11.66\%, and +4.17\% on average, respectively.
These observations highlight Fed-HeLLo's ability to adapt to severe data distribution challenges. The results indicate that Fed-HeLLo excels in more balanced scenarios while remaining effective in environments with significant resource and data heterogeneity, making it a robust solution for complex datasets like LEDGAR. {Regarding resource efficiency, our method reduces the average backward computational cost by 1.18 TFLOPs compared to rank-based methods under the 6:3:1 RoC scenario and by 0.78 TFLOPs under the 1:1:1 RoC scenario. It increases average memory usage by only 0.13 GB compared to Straggler Learning under the 6:3:1 RoC distribution. However, as the number of low-resource devices increases, this gap becomes negligible. Additionally, our method decreases training time by 0.34 seconds and 0.79 seconds compared to the random allocation method FedRA under the 6:3:1 and 1:1:1 RoC distributions, respectively, benefiting from the designed HLA strategy.
}

{\textbf{Quantitative Results on Natural Instruction:} 
We further evaluate performance on larger language models for instruction fine-tuning tasks, as presented in Table~\ref{tab:main-natural_instruction}. On Natural Instruction dataset with the 6:3:1 RoC distribution, our method achieves exceptionally high performance on IID scenario, outperforming FLoRA, Exclusive Learning, Straggler Learning, and HETLoRA by +18.03\%, +6.60\%, +6.47\%, and +5.27\%, respectively. However, our method outperforms FedRA by +5.27\% with our HLA design. While FlexLoRA performs reasonably well, our method still surpasses it by a considerable margin of +4.25\%. In the Non-IID scenario (15/1.0), our method shows a slight performance drop compared to FlexLoRA, as it may requires more training rounds to optimize LoRA layer allocation under Non-IID data distributions (see Section~\ref{sec:visualization_of_training} for further discussion).
Regarding computational efficiency, our method reduces the average backward computational cost by 1.41 TFLOPs compared to rank-based methods, with the reduction scaling linearly as batch size increases (set to 1 in our case). It also lowers average memory usage by 1.09 GB compared to rank-based methods and only increases it by 0.12 GB relative to Straggler Learning. In terms of communication cost, our method remains efficient since only the trainable LoRA layer’s parameters need to be uploaded, whereas FlexLoRA requires downloading the full-rank matrix of LoRA layers, resulting in significantly higher communication overhead. Additionally, our method strikes a balance between training time and model performance, increasing training time by only 1.52 seconds compared to Straggler Learning under the 6:3:1 distribution.
}

{\textbf{Quantitative Results on Dolly-15K:} 
Dolly-15K presents a real-world scenario with human-generated prompt-response pairs. As shown in Table~\ref{tab:main-dolly15k}, our method outperforms all compared approaches across all RoC distributions, consistently maintaining stable performance in both IID and Non-IID settings. Under the 6:3:1 RoC distribution, our method improves the average Rouge-L score by +1.89\%, +2.92\%, and +17.43\% compared to Straggler Learning, HETLoRA, and FLoRA, respectively. Despite using SVD to obtain LoRA layer parameters, FlexLoRA still falls behind, our method surpasses it by +0.77\% in performance. Our method achieves a +0.59\% improvement compared to FedRA.
The resource efficiency trends remain consistent under Dolly-15K dataset. Our method reduces the average backward computational cost by 0.67 TFLOPs and 0.34 TFLOPs compared to rank-based methods under the 6:3:1 RoC and 1:1:1 RoC distributions, respectively. It also lowers average memory usage by 1.09 GB and 0.57 GB for the two distributions. Additionally, our method maintains a low average communication cost compared to Exclusive Learning and FlexLoRA, while achieving a balanced trade-off in training time, requiring only 94.25 and 104.48 seconds per training round.
}

\textbf{Quantitative Results on DomainNet-121:} In DomainNet-121 experiments, as shown in Table~\ref{tab:main-domainnet121}, Fed-HeLLo further demonstrates its effectiveness, particularly in handling Non-IID data. Under the 6:3:1 RoC distribution, Fed-HeLLo achieves an average accuracy of 57.30\%, outperforming Straggler Learning by +19.60\% and Exclusive Learning by +22.25\%. In the 1:1:1 RoC setting, Fed-HeLLo continures to outperform baseline methods, with a +9.09\% improvement over HETLoRA in the ``25/1'' Non-IID scenario and a +6.31\% increase over Exclusive Learning. Even in the most challenging ``15/1'' Non-IID setting, Fed-HeLLo maintains its advantage, outperforming HETLoRA by +18.40\%. 
On average, Fed-HeLLo outperforms FedRA by +1.96\% and +1.83\% in the 6:3:1 and 1:1:1 RoC settings, respectively.
{FlexLoRA and FLoRA perform poorly on mid-sized model, showing significant performance drops across both RoC distributions.}
These results on the DomainNet-121 dataset reinforce Fed-HeLLo's capability to handle diverse and challenging FL environments, ensuring high performance even under extreme conditions.

\subsection{Visualization of the Training Process}\label{sec:visualization_of_training}
\begin{figure*}[t]
\centering
\includegraphics[width=0.9\linewidth]{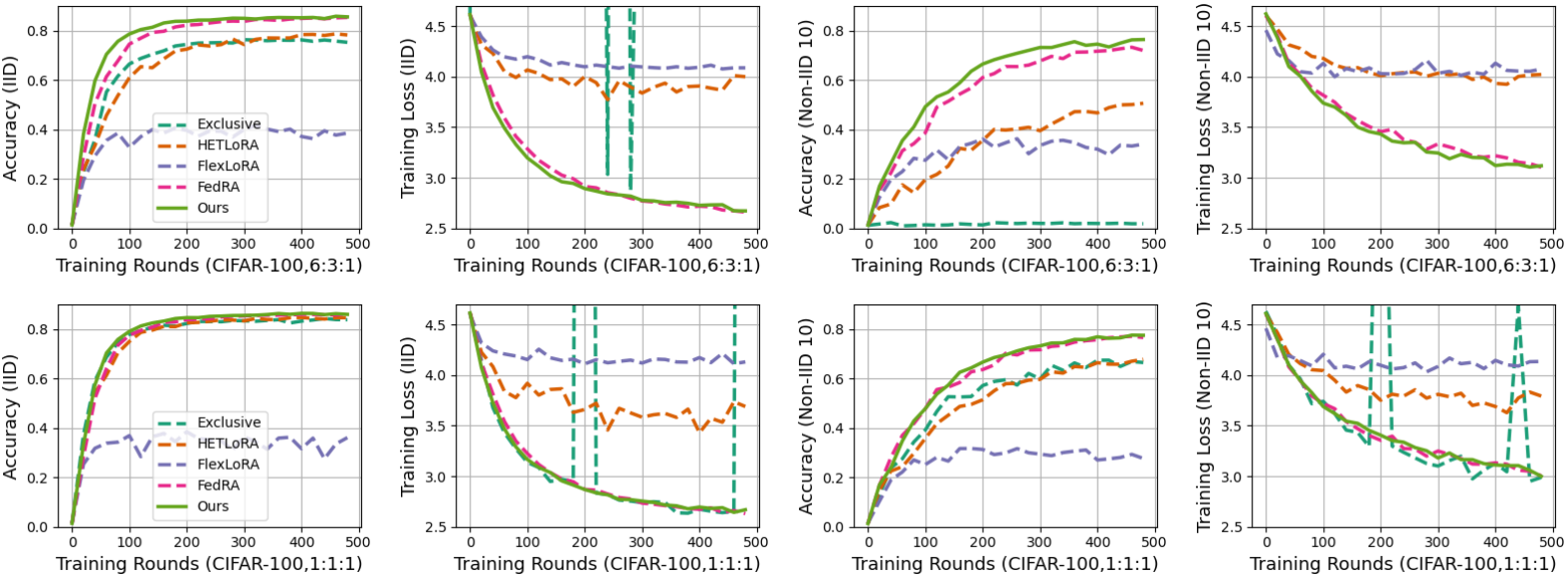}
\caption{{The accuracy and training loss with respect to training rounds on CIFAR-100 dataset, across different data and device settings.}}
\label{fig:acc-loss-cifar100}
\end{figure*}

{
We visualize the accuracy and loss curves during training with four methods (i.e., Exclusive Learning, HETLoRA, FlexLoRA, and FedRA) and our approach on CIFAR-100 dataset. As shown in Figure~\ref{fig:acc-loss-cifar100}, the first row presents results under the 6:3:1 RoC distribution, while the second row corresponds to the 1:1:1 RoC distribution. For each row, the first two plots illustrate accuracy and training loss under an IID data distribution, whereas the latter two depict results under a Non-IID setting.
Taking the accuracy and training loss curves under the 6:3:1 RoC distribution in an IID setting as an example, the plots show that our method (green solid line) achieves the highest accuracy and converges faster than other approaches. Notably, FlexLoRA (blue dashed line) exhibits significantly lower accuracy, indicating poor performance in this setting. FedRA (pink dashed line) and HETLoRA (orange dashed line) converge more slowly than our method, with HETLoRA stabilizing at a lower accuracy. The right plot illustrates training loss, where our method consistently reduces loss faster and reaches a lower loss, demonstrating better convergence. Exclusive Learning (deep green dashed line) shows instability with sudden fluctuations at the beginning of the training rounds, suggesting instability under this distribution. These results highlight the effectiveness of our approach in achieving both superior accuracy and stable training convergence compared to existing methods.
In the Non-IID setting of the 6:3:1 RoC distribution, our method converges slightly slower than FlexLoRA at the beginning, which aligns with the issue observed in the main results on the Natural Instruction dataset. Under the 1:1:1 RoC distribution, the performance gaps narrow slightly, but our method still outperforms other approaches, maintaining higher accuracy and better convergence stability.
} 

\subsection{Ablation Study}\label{exp:ablation-results}
\subsubsection{Cross-silo Setting Evaluation}

\begin{table}[t]
  \centering
  \caption{{Cross-silo evaluation on CIFAR-100 with 20 clients and two RoC settings.}}
  \scalebox{0.95}{
    \begin{tabular}{c|c|cccc}
    \toprule[1pt]
    \multicolumn{2}{c|}{\textbf{Dataset/Model}} & \multicolumn{4}{c}{\textbf{CIFAR-100/ViT-base}} \\
    \midrule    \multicolumn{2}{c|}{\textbf{Metric}} & \multicolumn{4}{c}{\textbf{Accuracy (\%)} $\uparrow$} \\
    \midrule
    \multicolumn{2}{c|}{\textbf{(Methods/Client Settings)/DH}} & \textbf{IID} & \textbf{20/1.0} & \textbf{10/1.0} & \textbf{Avg} \\
    \midrule
    Straggler & \multicolumn{1}{c|}{-} & 72.73 & 56.92 & \cellcolor[rgb]{ .749,  .749,  .749}46.01 & 58.55 \\
    \midrule
    Exclusive & \multicolumn{1}{c|}{\multirow{6}[0]{*}{\shortstack{RoC\\6:3:1}}} & 79.26 & 31.70 & 14.31 & 41.75 \\
    HETLoRA &       & 81.92 & 32.63 & 17.50 & 44.01 \\
    {FlexLoRA} & &{38.57}       &{15.06}  &{10.22}  &{21.27}   \\
    {FLoRA} &    &{2.12}  &{1.28}  &{1.62}  &{1.67}  \\
    FedRA &       & 86.19 & 68.03 & 3.00 & 52.40 \\
    \textbf{Fed-HeLLo} &       & \cellcolor[rgb]{ .749,  .749,  .749}86.63 & \cellcolor[rgb]{ .749,  .749,  .749}69.29 & 6.49  & \cellcolor[rgb]{ .749,  .749,  .749}54.14 \\
    \midrule
    Exclusive & \multicolumn{1}{c|}{\multirow{6}[0]{*}{\shortstack{RoC\\1:1:1}}} & 84.98 & 54.70 & 4.98  & 48.22 \\
    HETLoRA &       & 86.60 & 60.89 & 41.92 & 63.13 \\
    {FlexLoRA} &       &{45.51}  &{16.69}  &{2.44}  &{21.54}  \\
    {FLoRA} &       &{2.99}  &{1.68}  &{1.86}  &{2.17}  \\
    FedRA &       & 87.64 & 74.68 & 53.28 & 71.86 \\
    \textbf{Fed-HeLLo} &       & \cellcolor[rgb]{ .749,  .749,  .749}88.22 & \cellcolor[rgb]{ .749,  .749,  .749}75.09 & \cellcolor[rgb]{ .749,  .749,  .749}56.27 & \cellcolor[rgb]{ .749,  .749,  .749}73.19 \\
    \bottomrule[1pt]
    \end{tabular}%
    }
  \label{tab:abl-num-client}%
  \arrayrulecolor{black}
  
\end{table}%

Here, we conduct an ablation study to evaluate the stability of our proposed method under cross-silo client settings. 
Table~\ref{tab:abl-num-client} presents the cross-silo evaluation results on CIFAR-100 dataset using the ViT-base model, with 20 clients under two different RoC settings: 6:3:1 and 1:1:1. The table compares the performance of different methods across IID and two Non-IID scenarios (20/1.0 and 10/1.0). Under the 6:3:1 RoC setting, Fed-HeLLo significantly outperforms the other methods, achieving the highest average accuracy of 54.14\%. Specifically, Fed-HeLLo reaches 86.63\% accuracy in the IID case, outperform FedRA by +0.44\% and HETLoRA by +4.71\%. In the 20/1.0 Non-IID scenario, Fed-HeLLo achieves an accuracy of 69.29\%, which is +37.59\% higher than Exclusive Learning. However, all fix methods diverge in the ``10/1.0'' Non-IID case, showing low accuracies under 20\%, which may be a result from the imbalance of data among clients and heterogeneous client's model updates in this extreme Non-IID scenario. Under the 1:1:1 RoC setting, Fed-HeLLo continues to lead, achieving the highest average accuracy of 73.19\%. In the IID scenario, Fed-HeLLo reaches 88.22\% accuracy, which is +0.58\% better than FedRA and +3.24\% better than Exclusive Learning. In the 20/1.0 Non-IID case, Fed-HeLLo achieves an accuracy of 75.09\%, which is +0.41\% higher than FedRA and +14.20\% higher than HETLoRA. For the 10/1.0 Non-IID case, Fed-HeLLo's accuracy drops to 56.27\%, yet it outperforms FedRA by +2.99\% and HETLoRA by +14.35\%. {FlexLoRA and FLoRA perform even worse in cross-silo settings, demonstrating poor generalization to both the number of clients and mid-size FMs.}

\subsubsection{GD-HLA Evaluation}\label{subsubsection:GD-HLA}

\begin{table}[t]
  \centering
  \caption{GD-HLA evaluation on CIFAR-100 dataset with two client settings: 100 clients with a RoC of 6:3:1, and 20 clients with a RoC of 1:1:1.}
  \scalebox{0.9}{
    \begin{tabular}{c|c|c|cccc}
    \toprule[1pt]
    \multicolumn{3}{c|}{\textbf{Dataset/Model}} & \multicolumn{4}{c}{\textbf{CiFAR-100/ViT-base}} \\
    \midrule
    \multicolumn{3}{c|}{\textbf{Metric}} & \multicolumn{4}{c}{\textbf{Accuracy (\%)} $\uparrow$} \\
    \midrule
    \multicolumn{3}{c|}{\textbf{(Methods/Client Settings)/DH}} & \textbf{IID} & \textbf{20/1.0} & \textbf{10/1.0} & \textbf{Avg} \\
    \midrule
    Triangle ($\rhd$) & \multicolumn{1}{c|}{\multirow{4}[2]{*}{\shortstack{10/100\\Clients}}} & \multicolumn{1}{c|}{\multirow{4}[1]{*}{\shortstack{RoC\\6:3:1}}} & 84.02 & 53.51 & 54.18 & 63.90 \\
    In-Triangle ($\lhd$) &       &       & 78.50 & 69.03 & 58.47 & 68.66 \\
    Uniform ($\approx$) &       &       & 84.12 & 70.85 & 55.83 & 70.26 \\
    \textbf{Bottleneck ($\bowtie$)} &       &       & \cellcolor[rgb]{ .749,  .749,  .749}84.16 & \cellcolor[rgb]{ .749,  .749,  .749}78.19 & \cellcolor[rgb]{ .749,  .749,  .749}71.02 & \cellcolor[rgb]{ .749,  .749,  .749}77.79 \\
    \midrule
    Triangle ($\rhd$) & \multicolumn{1}{c|}{\multirow{4}[0]{*}{\shortstack{4/20\\Clients}}} & \multicolumn{1}{c|}{\multirow{4}[0]{*}{\shortstack{RoC\\1:1:1}}} & 87.29 & 64.75 & 6.30 & 52.78 \\
    In-Triangle ($\lhd$) &       &       & 85.76 & 66.64 & 2.34  & 51.58 \\
    Uniform ($\approx$) &       &       & 87.15 & 62.18 & 2.00 & 50.44 \\
    \textbf{Bottleneck ($\bowtie$)} &       &       & \cellcolor[rgb]{ .749,  .749,  .749}87.50 & \cellcolor[rgb]{ .749,  .749,  .749}70.54 & \cellcolor[rgb]{ .749,  .749,  .749}38.18 & \cellcolor[rgb]{ .749,  .749,  .749}65.41 \\
    \bottomrule[1pt]
    \end{tabular}%
    }
  \label{tab:abl-geo}%
\end{table}%

Table~\ref{tab:abl-geo} illustrates the performance evaluation of GD-HLA on CIFAR-100 dataset, comparing four HLA patterns: \textit{Triangle}, \textit{Inverted-Triangle}, \textit{Uniform}, and \textit{Bottleneck}, under two client settings: 100 clients with a 6:3:1 RoC and 20 clients with a 1:1:1 RoC. In the 100 client, 6:3:1 RoC scenario, the \textit{Bottleneck} HLA pattern consistently outperforms the others, achieving the highest average accuracy of 77.79\%. Specifically, \textit{Bottleneck} records an accuracy of 84.16\% in the IID setting, slightly surpassing \textit{Triangle} and \textit{Uniform}, which achieve 84.02\% and 84.12\% accuracy, respectively. In the Non-IID case (20/1.0), \textit{Bottleneck} again leads with 78.19\% accuracy, outperforming \textit{Uniform} by +7.34\%, \textit{In-Triangle} by +9.16\%, and \textit{Triangle} by +24.68\%. This trend continues in the 10/1.0 Non-IID case, where \textit{Bottleneck} achieves 71.02\% accuracy, outperforming \textit{Uniform} by +15.19\%, \textit{In-Triangle} by +12.55\%, and \textit{Triangle} by +16.84\%.
In the 20 client, 1:1:1 RoC scenario, \textit{Bottleneck} HLA continues to outperforms other GD-HLA strategies, achieving an average accuracy of 65.41\%. Specifically, \textit{Bottleneck} achieves an accuracy of 87.50\% in the IID scenario, slightly higher than \textit{Triangle} and \textit{Uniform}, which recorded 87.29\% and 87.15\%, respectively. However, in the Non-IID (20/1.0) scenario, \textit{Bottleneck's} accuracy drops to 70.54\%, yet it remains higher than \textit{Uniform} by +8.36\%, \textit{In-Triangle} by +3.90\%, and \textit{Triangle} by +5.79\%. For the extreme Non-IID (10/1.0) case, \textit{Bottleneck} records an accuracy of 38.18\%, outperforming the other patterns by a significant margin, indicating its effectiveness in handling high degrees of Non-IID data.

\subsubsection{The Dynamic and Intrinsic Layer Importance Evaluation}\label{subsubsec:intrinsic-dynamic-layer-importance}

\begin{figure}[t]
\centering
\includegraphics[width=0.9\linewidth]{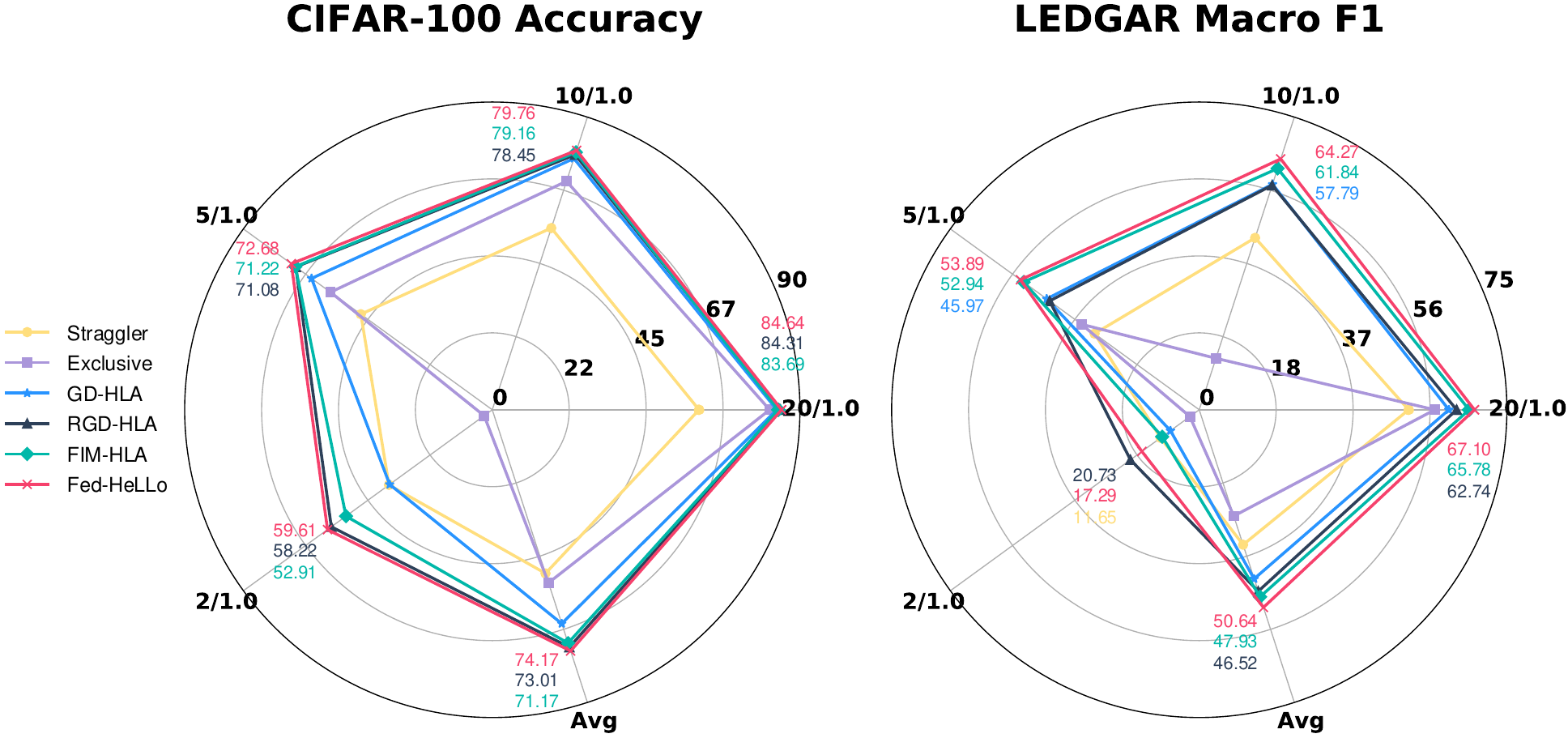}
\caption{The radar chart presents the results of the ablation study focused on evaluating the dynamic and intrinsic layer importance, specifically comparing our method with GD-HLA, RGD-HLA, FIM-HLA, and other baseline methods. The studies are conducted on both the CIFAR-100 and LEDGAR datasets, using 100 clients with a RoC distribution of 1:1:1. The Top-3 results for each spoke are listed near the vertices from top to bottom, with corresponding colors.}
\label{fig:importance-radar-chart}
\end{figure}

In designing the Fed-HeLLo framework, we demonstrate that incorporating both dynamic and intrinsic layer importance can enhance model performance. Figure~\ref{fig:importance-radar-chart} presents two radar charts comparing the performance of GD-HLA, RGD-HLA, FIM-HLA, two baseline methods (i.e., Straggler and Exclusive Learning), and our method, with each method represented by different colored lines. The evaluation is conducted on two datasets, CIFAR-100 and LEDGAR. The radial axes display various data distribution settings and overall average performance, with each point representing a specific data distribution scenario or an aggregated result.

In the chart on the left, Fed-HeLLo (represented by the red line) consistently achieves high accuracy across all Non-IID settings, particularly excelling in the ``5/1.0'' and ``10/1.0'' settings, where it reaches 72.68\% and 79.76\%, respectively. It also leads in the ``20/1.0'' setting, achieving 84.64\% in accuracy. Compared to other methods, Straggler Learning (yellow) performs the worst, especially in the ``20/1.0'' setting, where its accuracy falls below 50\%. Exclusive Learning (purple) and GD-HLA (blue) demonstrate moderate performance, with GD-HLA slightly outperforming Exclusive Learning in most settings. RGD-HLA (black) and FIM-HLA (green) perform well but lag behind Fed-HeLLo in the more extreme Non-IID scenarios. The overall average accuracy further underscores the superiority of Fed-HeLLo, as it consistently surpasses other methods across different data heterogeneity levels.

In the chart on the right, Fed-HeLLo again demonstrates strong performance across Non-IID settings, especially in the ``10/1.0'' and ``20/1.0'' scenarios, where it achieves 64.27\% and 67.10\%, respectively. Straggler Learning continues to perform poorly, particularly in the ``2/1.0'' setting, where it falls to 11.65\%. Exclusive Learning and GD-HLA show moderate improvements over Straggler Learning, but they still fall short of Fed-HeLLo’s performance. {RGD-HLA exhibits competitive performance, especially in the ``10/1.0'' setting, which indicates the stable training is important for the extreme Non-IID cases.} The average F1 scores indicate that Fed-HeLLo is not only robust but also adaptable to varying degrees of client and data heterogeneity, maintaining high F1 scores across the board.
Across both datasets, Fed-HeLLo consistently outperforms the other methods in nearly every Non-IID setting, as well as in the overall average metrics, demonstrating its effectiveness in FL with heterogeneous clients.

\begin{table}[t]
  \centering
  \caption{Justification of FIM-HLA with different training strategies.}
  \scalebox{1.05}{
    \begin{tabular}{c|cccc}
    \toprule[1pt]
    \textbf{Dataset/Model} & \multicolumn{4}{c}{\textbf{CIFAR-100/ViT-base}} \\
    \midrule
    \textbf{Metric} & \multicolumn{4}{c}{\textbf{Accuracy (\%)} $\uparrow$} \\
    \midrule
    \textbf{Methods/DH} & \textbf{IID} & \textbf{20/1.0} & \textbf{10/1.0} & \textbf{Avg} \\
    \midrule
    Proxy Data-1 & 84.02 & 77.27 & 69.95 & 77.08 \\
    Proxy Data-2 & 3.24  & 3.14  & 3.32  & 3.23 \\
    \textbf{Fed-HeLLo} & \cellcolor[rgb]{ .749,  .749,  .749}87.11 & \cellcolor[rgb]{ .749,  .749,  .749}84.64 & \cellcolor[rgb]{ .749,  .749,  .749}79.76 & \cellcolor[rgb]{ .749,  .749,  .749}79.03 \\
    \bottomrule[1pt]
    \end{tabular}%
    }
  \label{tab:abl-fim-1}%
\end{table}%

\begin{table}[t]
  \centering
  \caption{Justification of FIM-HLA with Wikipedia proxy dataset.}
  \scalebox{1}{
    \begin{tabular}{c|cccc}
    \toprule[1pt]
    \textbf{Dataset/Model} & \multicolumn{4}{c}{\textbf{DomainNet121/ViT-base}} \\
    \midrule
    \textbf{Metric} & \multicolumn{4}{c}{\textbf{Accuracy (\%)} $\uparrow$} \\
    \midrule
    \textbf{Methods/DH} & \textbf{IID} & \textbf{25/1} & \textbf{15/1} & \textbf{Avg} \\
    \midrule
    HETLoRA & 64.26 & 48.92 & 35.34 & 49.50 \\
    FedRA & 68.53 & 55.64 & 51.07 & 58.41 \\
    \textbf{Fed-HeLLo-Wikipedia} & \cellcolor[rgb]{ .851,  .851,  .851}68.65 & \cellcolor[rgb]{ .851,  .851,  .851}57.74 & \cellcolor[rgb]{ .851,  .851,  .851}52.94 & \cellcolor[rgb]{ .851,  .851,  .851}59.78 \\
    \textbf{Fed-HeLLo} & \cellcolor[rgb]{ .749,  .749,  .749}68.98 & \cellcolor[rgb]{ .749,  .749,  .749}58.01 & \cellcolor[rgb]{ .749,  .749,  .749}53.74 & \cellcolor[rgb]{ .749,  .749,  .749}60.24 \\
    \bottomrule[1pt]
    \end{tabular}%
    }
  \label{tab:abl-fim-2}%
\end{table}%

\subsubsection{Justification of FIM-HLA}\label{subsubsec:fim-hla}
\textbf{Training Strategy:}
In FIM-HLA, we assume that the server holds a small proxy dataset. Ideally, the server could directly use this proxy dataset to improve the global model performance, making FIM-HLA unnecessary. To demonstrate that a small proxy dataset alone cannot significantly enhance fine-tuning performance and to justify the need for FIM-HLA, we conduct two additional experiments on CIFAR-100 dataset, named \textit{Proxy Data-1} and \textit{Proxy Data-2}. All experiments sample 10 out of 100 clients with a RoC of 1:1:1. In the first experiment, we implement Fed-HeLLo but with the server acting as an additional client with sufficient computational capability to fine-tune all LoRA layers using the proxy dataset. In the second experiment, we implement Fed-HeLLo but with the server using the proxy data to fine-tune the global model after aggregation at each round. The accuracy results for both experiments are reported in Table~\ref{tab:abl-fim-1}. 
We observe that both proxy data-based global model fine-tuning methods result in unfavorable performance compared to FIM-HLA. 
For \textit{Proxy Data-1}, there are consistent performance drops compared to \textit{Fed-HeLLo}. This may be due to the server updates trained on proxy data being included in the model updates each round, causing the model to overfit to the proxy data and leading to suboptimal performance. 
For \textit{Proxy Data-2}, where the server fine-tunes the global model using proxy data after each aggregation round, the overfitting issue becomes even more severe, resulting in poor performance across all data distributions.
\textbf{Proxy Dataset from Wikipedia:} To further assess the sensitivity of the proxy data, we manually collect a dataset from Wikipedia\footnote{\url{https://www.wikipedia.org/}} for DomainNet-121 dataset. Specifically, we use each category name as a keyword, search for it on Wikipedia, and download the first image that appears, resulting in a small proxy dataset of 121 images (one per category). In Table~\ref{tab:abl-fim-2}, we present the experimental results using a 10/100 clients setting with a RoC of 1:1:1. For Fed-HeLLo, we use 1,000 random test data samples from DomainNet-121, which are excluded from all other training and evaluation. For Fed-HeLLo-Wikipedia, we utilize the proxy dataset collected from Wikipedia, following the same training conditions as Fed-HeLLo. In terms of results, Fed-HeLLo consistently outperforms all other methods across all scenarios, achieving the highest accuracy in the IID (68.98\%), 25/1 Non-IID (58.01\%), and 15/1 Non-IID (53.74\%) settings, with an overall average accuracy of 60.24\%. Fed-HeLLo-Wikipedia closely follows, particularly in the IID (68.65\%) and 25/1 Non-IID (57.74\%) settings, but with a slightly lower average accuracy of 59.78\%. This drop is possibly due to the proxy data from Wikipedia lacking domain diversity like DomainNet-121 and having fewer samples. FedRA falls behind in all scenarios, with an average accuracy of 58.41\%, while HETLoRA performs the worst, particularly in Non-IID scenarios, leading to the lowest overall average accuracy of 49.50\%.

\section{Limitations and Potential Future Works}\label{sec:limitation}
{\textbf{Privacy Preserving.} While our work leverages the inherent privacy-preserving nature of FL by ensuring that client data remains local and is not shared with the server or other clients, potential vulnerabilities still exist. For example, FL remains susceptible to privacy attacks such as model inversion~\cite{fredrikson2015model}, where adversaries attempt to reconstruct client data from model updates. Techniques like differential privacy~\cite{dwork2006differential} and secure aggregation~\cite{bonawitz2016practical} can be integrated into our framework to mitigate the vulnerability.}

{\textbf{Memory Constraint.} Additionally, the storage of intermediate activations is still challenging to resource-constrained clients. Activation saving techniques, such as sparsification~\cite{rao2021dynamicvit} and pruning~\cite{zhu2017prune}, can be integrated to reduce storage overhead while maintaining training stability. Future work can explore adaptive activation compression strategies to dynamically adjust compression levels based on available client resources, minimizing memory consumption without significantly impacting model performance.}

{\textbf{Training Process Improvement.} Furthermore, while Fed-HeLLo introduces HLA to reduce memory overhead, a better training process can be designed to improve the FL process. For example, model distillation~\cite{li2019fedmd} can be used to obtain a smaller FM for clients to reduce the storage burden.}

{\textbf{Data Heterogeneity.} While Fed-HeLLo effectively optimizes resource allocation through HLA, it does not directly address the challenge of data heterogeneity. 
A potential future direction to mitigate the effects of data heterogeneity is to incorporate techniques such as federated feature alignment~\cite{yu2021fed2}, and adaptive weighting strategies~\cite{wu2021fast}.}

{By addressing these challenges, Fed-HeLLo can be further improved to enhance scalability, efficiency, and applicability in real-world FL scenarios.}

\section{Conclusion}
In this paper, we introduce Fed-HeLLo, a novel federated LoRA-based fine-tuning framework that integrates various HLA strategies.
These strategies allocate trainable LoRA layers based on clients' resource capabilities while considering both dynamic and intrinsic layer importance. 
To capture dynamic layer importance, we leverage FIM scores and design the FIM-HLA strategy. 
Additionally, to stabilize training in the early stages, 
we propose GD-HLA and RGD-HLA strategies that account for intrinsic layer importance. By co-designing these strategies, Fed-HeLLo incorporates a comprehensive HLA design.
Extensive experiments on five LoRA-based fine-tuning datasets validate the effectiveness and efficiency of Fed-HeLLo, offering valuable insights for designing the HLA in FL with heterogeneous resource clients.


{
\small
\bibliographystyle{plain}
\bibliography{ref.bib}
}

\vfill

\end{document}